\documentclass[11pt,a4paper]{article}
\usepackage{times,latexsym}
\usepackage{url}
\usepackage[T1]{fontenc}

\usepackage[acceptedWithA]{tacl2021v1}

\usepackage{amsmath}
\usepackage{amssymb}
\usepackage{amsthm}
\usepackage{graphicx}
\usepackage{booktabs}
\usepackage{multirow}
\usepackage{algorithm}
\usepackage{algorithmic}
\usepackage{enumitem}

\newcommand{\surprisal}{\mathcal{S}}
\newcommand{\entropy}{\mathcal{H}}
\newcommand{\prob}{\mathbb{P}}
\newcommand{\probA}{\mathbb{P}_A}

\DeclareMathOperator*{\argmin}{arg\,min}

\newtheorem{definition}{Definition}

\title{Extending Minimal Pairs with Ordinal Surprisal Curves and Entropy Across Applied Domains}

\author{
  Andrew Katz \\
  Department of Engineering Education \\
  Virginia Tech \\
  \texttt{akatz4@vt.edu}
}

\date{}

\begin{document}

\maketitle

\begin{abstract}
The minimal pairs paradigm of comparing model probabilities for contrasting completions has proven useful for evaluating linguistic knowledge in language models, yet its application has largely been confined to binary grammaticality judgments over syntactic phenomena. Additionally, standard prompting-based evaluation requires expensive text generation, may elicit post-hoc rationalizations rather than model judgments, and discards information about model uncertainty. We address both limitations by extending surprisal-based evaluation from binary grammaticality contrasts to ordinal-scaled classification and scoring tasks across multiple domains. Rather than asking models to generate answers, we measure the information-theoretic ``surprise'' (negative log probability) they assign to each position on rating scales (e.g., 1--5 or 1--9), yielding full surprisal curves that reveal both the model's preferred response and its uncertainty via entropy. We explore this framework across four domains: social-ecological-technological systems classification, causal statement identification (binary and scaled), figurative language detection, and deductive qualitative coding. Across these domains, surprisal curves produce interpretable classification signals with clear minima near expected ordinal scale positions, and entropy over the completion tended to distinguish genuinely ambiguous items from easier items. 

\end{abstract}

\section{Introduction}
\label{sec:introduction}

Large language models (LLMs) are increasingly used for tasks such as classification, assessment, and decision-making across diverse domains \citep{brown2020language, chowdhery2022palm, openai2023gpt4}. There are suggestions that these models demonstrate emergent capabilities \citep{wei2022emergent,berti2025emergent} and can learn representations of the world \citep{yildirim2024task}. Myriad evaluation paradigms exist, yet such approaches face tradeoffs. For example, explicit prompting requires API access to any model and then comparing the generated output with ground truth answers, but it can be expensive for text generation and that generated reasoning \citep{wei2022chain, kojima2022large} may constitute post-hoc rationalizations rather than genuine understanding \citep{turpin2023language,lanham2023measuring}. Alternatively, binary outputs are very simple but lack the uncertainty quantification necessary for high-stakes applications \citep{guo2017calibration}. If asked directly, a model may be very confident in its answer, but that confidence may not be well-calibrated \citep{geng2024survey}. Finally, models may encode knowledge in their learned representations that they do not readily articulate through explicit generation. These limitations motivate the search for complementary evaluation approaches that can efficiently access models' representations and uncertainties in a quantifiable way without depending on text generation alone. Our goal is to find a method to probe LLM representations by measuring their surprisal at encountering different tokens as a means to (a) glean insight into those representations and (b) identify how LLMs could be used in downstream tasks using this mechanism.

\section{Background and Related Work}
\label{sec:related}

\subsection{Building on a Simple Idea: Surprisal as a Window into Models' Representations}

The key idea motivating the work here is simple. We tend to form expectations about the world based in part on our internal representations of our surroundings \citep{hohwy2020new,sprevak2023introduction,pezzulo2022evolution}. When we encounter something unexpected, by definition, that is something for which we assigned a low probability. Likewise, LLMs may also form internal representations of the world, which then inform their expectations over tokens they would encounter in text. Current autoregressive LLMs (by design) generate probability distributions of tokens conditioned on prior contexts. Logically, that means some tokens would be more likely to be in a generated text completion compared with other tokens, given the preceding tokens. For example, if a model with sufficient world knowledge is asked to complete the statement ``The capital of France is\dots'' then the next token is more likely to be ``Paris'' than ``Tokyo'' because the model has learned that ``Paris'' is the capital of France and not ``Tokyo'' (or at least that it appears at the end of such statements more, based on its training data). Similarly, if the statement ``The capital of France is now'' concluded with ``Tokyo'' then that would be an unlikely completion and be more \textit{surprising} to the model (and readers!). The basic hypothesis is that LLMs may have internal representations given their training and contexts, and those representations lead to the models being more or less surprised by certain completion tokens. Thus, if we wanted an indirect way to probe those representations, then we could measure the model's surprise at encountering different tokens. NB: when we say ``the model is surprised'' we mean that in the information-theoretic sense and not in an anthropomorphization sense.

The information-theoretic measure that captures this intuition is \emph{surprisal}. Mathematically, surprisal is the negative log probability of an event, quantifying its unexpectedness. Lower surprisal corresponds to higher probability events (or tokens) occurring, reflecting what a person (or the model) considers more ``natural'' or ``expected.'' In the extreme, for certain events where $p(x) = 1$ then the surprisal is defined as 0. That should make sense because finding out that something is true when you already know it is true is not surprising. Conversely, if $p(x) = 0.001$ then the surprisal associated with finding out $x$ occurred is very large because an improbable event actually occurring is very surprising. Building an evaluation framework on that idea hypothetically may provide direct access to models' learned probability distributions and representations without requiring text generation. Thus, rather than asking an LLM a question like ``What is the answer?'' and analyzing its generated text, we can measure surprisal for alternative completions: ``The answer is X'' versus ``The answer is Y.'' This is the premise behind the minimal pairs evaluation paradigm \citep{zhou2025linguistic}. 

% \subsection{Surprisal in Cognitive Science}

Surprisal has deep connections to human language processing. Psycholinguistic research suggests that in some cases word surprisal correlates with reading times \citep{hale2001probabilistic, levy2008expectation, smith2013effect}, eye-tracking measures \citep{demberg2008data}, and neural responses (N400 ERP component) \citep{frank2015erp, kuperberg2016we}, though this is still a matter of debate \citep{slaats2025s}. \citet{hale2001probabilistic} proposed that processing difficulty is proportional to surprisal, formalizing the intuition that unexpected words are harder to process. \citet{levy2008expectation} extended this to the \emph{surprisal theory} of sentence processing, which has been extensively studied as a popular theory across languages and constructions \citep{smith2013effect}. This suggests that some aspects of human language comprehension may involve predictive processing, where the brain continuously generates expectations about upcoming input \citep{williams2018predictive}. Our work leverages these insights: if surprisal reflects processing difficulty for humans, it may similarly reveal what LLMs find ``natural'' or ``expected,'' providing a window into their learned representations.

\subsection{Formal Information Theory Foundations}

% \subsubsection{Shannon Entropy and Surprisal}

Information theory, introduced by \citet{shannon1948mathematical}, provides a mathematical foundation for quantifying information content and uncertainty in random variables, signals, or distributions \citep{lombardi2016shannon}. In that framework, the \emph{surprisal} (or self-information) of an event $x$ is defined as:

\begin{equation}
\surprisal(x) = \log (\frac{1}{\prob(x)}) = -\log \prob(x)
\end{equation}

Surprisal quantifies the ``unexpectedness'' of observing $x$. Rare events have high surprisal while common events have low surprisal. Following from that definition, the \emph{entropy} of a discrete probability distribution $X$ measures average surprisal:

\begin{equation}
\entropy(X) = -\sum_{x \in X} \prob(x) \log \prob(x)
\end{equation}

Entropy quantifies uncertainty, which means that uniform distributions have higher entropy while peaked distributions have lower entropy.

In language modeling, surprisal measures how unexpected a token is given its context \citep{jurafsky2000speech}. For a token $t_n$ following context in the window $t_{1:n-1}$:

\begin{equation}
\surprisal(t_n | t_{1:n-1}) = -\log \prob(t_n | t_{1:n-1})
\end{equation}

Average surprisal across a corpus relates to perplexity, a standard language model quality metric \citep{jelinek1977perplexity}.

\subsection{Related Approaches and Prior Work}

A prevalent evaluation paradigm involves prompting models to generate answers, often with reasoning chains \citep{wei2022chain, kojima2022large}. Large-scale benchmarks like HELM \citep{liang2022holistic}, GPQA \citep{rein2024gpqa}, and BIG-bench \citep{srivastava2022beyond} evaluate models across diverse tasks using prompting. When the benchmark involves multiple-choice questions, accuracy is easily computed, but if the task is more open-ended then even another LLM may be used for evaluation \citep{gera2025justrank,jiang2025codejudgebench,wang2025can}. While effective, prompting-based evaluation has limitations. \citet{turpin2023language} show that generated explanations may mislead humans, with models producing plausible-sounding reasoning that didn't reflect their actual decision process. \citet{lanham2023measuring} found that models can generate convincing arguments for incorrect answers. Additionally, text generation is computationally expensive, limiting evaluation scale.

An alternative approach to token output-based evaluation trains \emph{probing classifiers} on model representations to test whether specific information is encoded \citep{belinkov2017neural, conneau2018you, manigrasso2024probing}. Linear probes on hidden states can detect information about syntax \citep{hewitt2019structural}, semantics \citep{tenney2019bert}, and facts \citep{petroni2019language}. However, probing has its own limitations. \citet{rogers2020primer} noted that probes require training data and may not reflect how models actually use encoded information. \citet{pimentel2020information} further argued that probe performance depends on representation geometry, not just information content.

\subsubsection{Prior Work on Surprisal-Based Analysis}

The minimal pairs methodology entails comparing model behavior on carefully constructed sentence pairs that differ in a single linguistic feature. \citet{marvin2018targeted} introduced this approach for systematic language model evaluation, testing models on subject-verb agreement and other syntactic phenomena. The standard paradigm measures whether a model assigns higher probability to the grammatical sentence than its ungrammatical counterpart. This approach has also proven valuable for testing humans on reading tasks \citep{sprouse2013comparison}. \citet{wilcox2018rnn, wilcox2019structural} extended minimal pairs evaluation to study how models represent syntactic structure, finding that surprisal patterns can reveal hierarchical representations even in models with recurrent architectures. \citet{warstadt2020blimp} further scaled the approach by creating BLiMP (Benchmark of Linguistic Minimal Pairs), a benchmark comprising 67 datasets with 67,000 minimal pairs targeting grammaticality judgments across morphological, syntactic, and semantic phenomena. More recently, \citet{hu2024language} demonstrated that LLM probability judgments align with human grammaticality intuitions across a range of constructions.

Beyond minimal pairs, surprisal has been used to evaluate masked language models \citep{salazar2020masked} and analyze syntactic knowledge \citep{misra2020exploring, wilcox2020targeted}. These works demonstrated that probability-based evaluation is methodologically promising for assessing aspects of linguistic knowledge. However, they focused primarily on grammatical phenomena like syntax, morphology, and basic semantics. Our work extends this paradigm in two directions: first, from binary grammaticality judgments to ordinal scales that capture degrees of confidence; second, from linguistic phenomena to classification tasks across diverse applied domains.

Another critical methodological consideration for probability-based evaluation is surface form competition. This phenomenon refers to the observation that token probabilities are influenced by surface-level features independent of semantic content. \citet{holtzman2021surface} demonstrated that the highest-probability answer is not always semantically correct because probability mass is distributed across surface variants (e.g., ``Paris'', `` Paris'', ``paris'', ``PARIS''). This means that naive probability comparisons can yield misleading results. We address surface form competition through careful prompt design: ensuring consistent formatting, using leading spaces to match natural tokenization, and testing multiple response formats as robustness checks. Additionally, \citet{zhao2021calibrate} proposed contextual calibration for few-shot learning, estimating and correcting for content-free prior biases. While we do not apply explicit calibration in the present work, the principle of being aware of surface-level biases informs our experimental design.

\subsubsection{LLM Calibration and Confidence}

Recent work has also examined whether LLMs can reliably express confidence in their outputs. \citet{kadavath2022language} found that models can partially self-assess their knowledge, producing higher confidence for correct answers, but this self-assessment is imperfect. \citet{geng2024survey} provided a comprehensive survey of confidence estimation and calibration in LLMs, distinguishing between verbalized confidence (asking models to state their confidence), consistency-based methods (measuring agreement across samples), and probability-based methods (our approach). These studies suggest that LLMs still tend to be poorly calibrated when reporting their own confidence. As such, probability-based confidence measures have a theoretical advantage over the LLM's reported confidence: they directly access the model's learned distribution instead of relying on the model's ability to accurately report its own uncertainty. \citet{xiong2023can} contrasted verbalized uncertainty with actual probability distributions, finding that expressed confidence often diverges from underlying probabilities. Our entropy-based uncertainty quantification offers a principled alternative aligned with those findings that does not require models to articulate their confidence.

\subsection{Contributions of This Work}

Despite the substantial literature at the intersection of LLM evaluation and capabilities assessments, a gap remains: very few unified frameworks exist for applying surprisal-based evaluation to classification tasks beyond grammaticality. Existing work has explored this idea of surprisal-based LLM evaluation via minimal pairs analysis \citep{leivada2025large,pistotti2025exploring,sinha2023language,park2021deep}; the calibration literature emphasizes verbalized confidence or sampling-based methods; and few, if any, systematic studies have examined how context affects surprisal-based classification across diverse domains. To be sure, log-probability scoring for multiple-choice tasks is already used in practice. Examples include \citet{brown2020language} selecting answers based on completion likelihood in GPT-3 evaluations, benchmarks such as MMLU \citep{hendrycks2020measuring} offer log-prob scoring as an evaluation mode, and the commonly used \texttt{lm-evaluation-harness} framework \citep{gao2024framework} implements log-likelihood-based scoring as a standard evaluation method across hundreds of benchmarks. However, these applications tend to use log-probabilities as a scoring convenience rather than as a formalized evaluation framework. Moreover, these approaches tend not to extend beyond binary or multiple-choice answer selection to ordinal scales or systematic uncertainty quantification through entropy.

We address this gap by extending the minimal pairs paradigm in two directions. First, we move from binary grammaticality judgments to ordinal scales (1-5, 1-9) that capture degrees of confidence and enable richer uncertainty quantification through entropy. Second, we apply this extended paradigm to practical classification tasks such as causal reasoning, figurative language detection, entity classification, and qualitative data labeling. Doing so helps demonstrate the approach's generality beyond linguistic acceptability. As such, this framework may be particularly valuable for large-scale evaluation, uncertainty-aware applications, and understanding implicit model knowledge.

\section{Methodology: Surprisal-Based Evaluation Framework}
\label{sec:methodology}

\subsection{Core Formulation}

\subsubsection{Surprisal for Classification and Scoring} 

Consider a classification task where you are observing someone else being asked to identify whether a ``vegetable'' is an instance of animal or a plant. You might be more surprised to hear that person say that ``vegetable'' is an animal than to hear that it is a plant. That is probably because you have some existing knowledge about the world that then tempers your expectations of what you will hear. That `tempering' is just conditioning a probability distribution given information about the world. In other words, you are computing $p(\text{``vegetable'' is a plant} | \text{world knowledge})$ and $p(\text{``vegetable'' is an animal} | \text{world knowledge})$. Based on those probabilities, you then form expectations of what you will hear, leading to that inverse relationship between surprisal and probability. Most current LLMs produce probability distributions over tokens. From that formulation, we can leverage this property to evaluate their performance on a wide range of tasks. A typical minimal pairs experiment would then compare the surprisal of the two options, leading to a classification of which option is more likely (or less surprising) given the context. These are commonly performed in grammar tasks \citep{he2024decoding,hu2024language}, but we extend the idea to classification tasks. That is, instead of asking which of two competing completions is more or less surprising, we ask which of a range of options on a scale is more or less surprising. 

For our approach, we focus on classification tasks where we want to determine a model's preference among alternative completions of a statement $\{a_1, a_2, \ldots, a_n\}$ given context $c$. Traditional prompting might ask the model to generate an answer. Instead, we construct prompts with contexts that end just before the target token and measure surprisal for each alternative, often in the form of a 5-point or 9-point scale. Extending this to ordinal scales, consider the scenario where you are taking a survey and are presented with the question ``On a scale from 1-5, where 1 means `strongly disagree'` and 5 means `strongly agree', how strongly do you agree with the following statement: Sunsets are beautiful.''. You might be asked to respond with a number from 1 to 5, and you might be surprised to see someone respond with a ``1'' given what you know about many people enjoying sunsets. Moreover, you might be less surprised to see someone with a rating of ``2'' compared to a ``1'' and so on. You have learned something about the world and human preferences that leads you to expect people responding with a higher number on that 1-5 preference scale. We leverage the same idea here and frame the task as a survey question for the LLM to respond to. The prompt sets up the task and rating scale as part of the context and ends just before the target token, which in this case would be the number on that scale. We then measure surprisal for each scale position (e.g., ``1'', ``2'', ``3'', ``4'', ``5''). In theory, the position with minimum surprisal represents the model's more ``natural'' or ``expected'' response. Additionally, by measuring the surprisal for each position, we can also quantify the model's uncertainty or confidence in its response. A steeper drop in surprisal across the scale positions might indicate lower uncertainty (viz. entropy) in the LLM, while a more gradual drop indicates higher uncertainty (i.e., higher entropy). Visually, this would look like a spikier surprisal curve or a smoother surprisal curve, respectively. Mathematically, this is expressed as follows.

\begin{definition}[Completion-Based Surprisal]
Given context $c$ and a set of alternative completions $A = \{a_1, \ldots, a_n\}$, the surprisal of alternative $a_i$ is:
\begin{equation}
\surprisal(a_i | c) = -\log \prob(a_i | c)
\end{equation}
where $\prob(a_i | c)$ is the model's probability of generating token $a_i$ given context $c$.
\end{definition}

The alternative completion $a$ with minimum surprisal, $a^*$, represents the model's most ``natural'' or ``expected'' completion:

\begin{equation}
a^* = \argmin_{a_i \in A} \surprisal(a_i | c)
\end{equation}

The central idea is that identifying $a^*$ for a model can reveal something about the model's learned representations and understanding of the task. This applies across task types ranging from binary classification to ordinal scoring, as described below.

\subsubsection{Task-Specific Applications}

\paragraph{Binary Classification.} For binary tasks (e.g., identifying a statement as expressing a causal relationship vs. non-causal relationship), we compare surprisal for two completions, similar to typical minimal pairs used in psycholinguistics:
\begin{equation}
\text{Class.} = \begin{cases}
\text{Pos.} & \text{if } \surprisal(\text{``T''} | c) < \surprisal(\text{``F''} | c) \\
\text{Neg.} & \text{otherwise}
\end{cases}
\end{equation}
where ``T'' and ``F'' denote ``True'' and ``False'' completions, respectively. The completion with minimum surprisal is the one the model deems most likely given the context. The surprisal difference $\Delta\surprisal = |\surprisal(a_1 | c) - \surprisal(a_2 | c)|$ may also provide a confidence measure, as larger differences indicate stronger model preference, though the extent to which this relationship holds remains an open empirical question.

\paragraph{Ordinal Scoring.} Moving beyond binary choices, we can frame tasks on ordinal scales (e.g., 1--5 or 1--9) where scale positions are not independent classes but are related through their ordering. This is our key extension of the minimal pairs paradigm. For an $n$-point ordinal scale, we measure surprisal for each possible position and identify the model's preferred score:
\begin{equation}
\text{Score}^* = \argmin_{s \in \{1, \ldots, n\}} \surprisal(s | c)
\end{equation}

The resulting \emph{surprisal curve} across all scale positions provides richer information than a single classification decision. Moreover, different anchor wordings yield different task framings. For example, we can apply a bipolar scale (1 = ``strongly disagree,'' 5 = ``strongly agree'') versus a unipolar scale (1 = ``very low,'' 5 = ``very high'') and then we can test multiple framings for robustness. The choice of scale length is itself an additional design variable: expanding from 1--5 to 1--9 may allow finer-grained evaluation, though the tradeoff between scale granularity and measurement quality is yet another open question.

\subsubsection{Uncertainty Quantification}

A key advantage of surprisal-based evaluation (compared to question-and-answer approaches) is uncertainty quantification through entropy. Although we could ask the model to report its own confidence or uncertainty, this may require additional training and calibration, neither of which can we assume are embedded in the model. However, through measuring surprisal over a range of possible ordinal completions, we can derive a probability distribution over the possible completions and then use that for an entropy calculation. An important detail is how this conversion is performed. The model produces logits (unnormalized log-probabilities, from which we compute probabilities and thus surprisal) over its full vocabulary $V$ at the target position. We restrict attention to only the tokens corresponding to our predefined alternatives $A = {a_1, \ldots, a_n}$ and renormalize over this restricted set. We denote these renormalized probabilities $\probA$ to distinguish them from the raw model probabilities $\prob$ used in Definition~1:

\begin{equation}
\probA(a_i | c) = \frac{\exp(\text{logit}(a_i | c))}{\sum_{j=1}^{n} \exp(\text{logit}(a_j | c))}
\label{eq:normalization}
\end{equation}

This renormalization ensures that $\sum_{i=1}^{n} \probA(a_i | c) = 1$, yielding a valid probability distribution over the alternatives of interest. Note that this discards probability mass assigned to tokens outside $A$ ($V \setminus A$); in other words, we condition on the assumption that the model's response is one of the predefined alternatives. This is analogous to a forced-choice experimental paradigm. Because renormalization preserves the ordering of probabilities, the classification decisions in Equations~5--7 (which depend only on the argmin) are identical under either $\prob$ or $\probA$.

With these normalized probabilities, the entropy of the distribution over alternatives, which quantifies model uncertainty, can then be calculated as:

\begin{equation}
\entropy(A | c) = -\sum_{i=1}^{n} \probA(a_i | c) \log \probA(a_i | c)
\end{equation}

Where $\probA(a_i | c)$ is the renormalized probability of completion $a_i$ given context $c$ as defined in Equation~\ref{eq:normalization}. High entropy indicates the model is uncertain among alternatives; low entropy indicates a strong preference. This provides a candidate principled confidence measure without requiring calibration on held-out data or model self-report, and may serve as a useful signal for downstream applications (this claim requires further empirical validation.) A further practical advantage is computational efficiency: surprisal-based evaluation requires only a single forward pass, reading out logits for a small set of tokens. For a binary classification task with short chain-of-thought reasoning, explicit prompting might generate 50--100 tokens of reasoning, while surprisal measurement requires computing logits for just 2 tokens, yielding a speedup in evaluations.

\subsubsection{Experimental Design Considerations}

The prior calculations were all conditioned on a fixed context $c$. This opens research avenues for exploring the impact of context and framing on model surprisal. Our experiments vary context levels to study how information provision affects surprisal patterns (see Section~\ref{sec:experiments} for details). Some experimental design principles and patterns include prompt structure and context manipulation. For example, each prompt combines relevant context, a task description, and an incomplete statement positioned so the model completes with the target token. Likewise, for context manipulation, we can test a gradient of content levels such as no context, a brief definition, or a comprehensive background of the task, all with the goal of studying how information provision affects surprisal patterns.

With this kind of setup, it is straightforward to employ factorial experimental designs to systematically study multiple factors. The \emph{model factor} involves testing multiple LLMs for cross-model validation. This can be developed for models within the same architecture (i.e., models of different sizes), or across different architectures (e.g., Qwen vs. LLaMA vs. Phi vs. etc.). \emph{prompt factors} include variations in personas, context levels (e.g., no context vs. brief definition of tasks vs. comprehensive background), and section delimiters (e.g., XML vs. Markdown vs. none). Varying those prompt factors allows us to study how different prompt designs affect surprisal patterns. Finally, \emph{task factors} capture domain-specific variables such as statement types and difficulty levels. If we can associate the tasks with difficulty ratings from human raters, then we can study how task difficulty affects surprisal patterns. This factorial structure enables analysis of main effects and interactions, providing robust insights into what factors influence surprisal-based performance.

For scoring tasks, we can analyze surprisal curves to extract multiple insights. The minimum location identifies the optimal score, that is, the scale position with the lowest surprisal. In theory, the steepness of the curve around that minimum reflects the model's confidence: steep curves indicate high confidence, whereas flat curves suggest low confidence. Multi-modality, the presence of multiple local minima, can signal genuine ambiguity in the task or competing interpretations within the model. Finally, asymmetry in the curve, where one tail is steeper than the other, may reveal directional biases in the model's judgments. We briefly validate surprisal-based classifications through comparisons with human judgments on some experiments below. Systematic calibration studies across domains and models remain an important direction for future work.

\section{Experiments and Results}
\label{sec:experiments}

We demonstrate our surprisal-based evaluation framework across four domains, each following the methodology in Section~\ref{sec:methodology}. While these domains may appear disparate, they share a common origin: our research program on \textit{mental models of social-ecological-technological systems} \citep{jones2011mental,rouse1986looking}, which studies how people understand complex systems through the entities they identify, the causal relationships they describe, the figurative language they use, and the thematic codes that characterize their responses. Each domain thus represents a task where LLM-assisted analysis could support qualitative research, and where surprisal-based evaluation offers a principled way to assess LLM performance. Table \ref{tab:experiments} summarizes the domains and task types.

\begin{table}[t]
\centering
\small
\footnotesize
\begin{tabular}{@{}p{2.5cm}ll@{}}
\toprule
\textbf{Domain} & \textbf{Task Type} & \textbf{Scale} \\
\midrule
SETS Classification & Score Discovery & 1-9 \\
Causal Statement & Scaled Classif. & 1-5, 1-9 \\
Figurative Lang. & Multi-format & Binary + Scales \\
Deductive Coding & Code Applic. & 1-5, 1-9 \\
\bottomrule
\end{tabular}
\caption{Summary of experimental domains and task types.}
\label{tab:experiments}
\end{table}

\subsection{Social-Ecological-Technological Systems Scoring}

\subsubsection{Task Description}

The Social-Ecological-Technological Systems (SETS) framework \citep{mcphearson2022social} analyzes entities across three interconnected dimensions: social, ecological, and technological. In their foundational paper, the authors displayed a ternary plot showing entities arranged somewhere in the two-dimensional triangle along the three social-ecological-technological dimensions. In this first task, we tested the extent to which LLMs would differentiate entities based on their social-ecological-technological dimensions by testing whether the surprisal-based approach can identify appropriate scores for entities on these dimensions.

\subsubsection{Methodology}

For each entity (e.g., ``park,'' ``current,'' ``web'') from a set of statements, for  each dimension (i.e., social, ecological, technological), we measure surprisal for score assignments on a 1-9 scale. The statements were crafted so that pairs of sentences with homonyms (e.g., web, current, spring) would have different meanings given the context. A representative prompt is shown below; the full set of prompt variations is described in Appendix~\ref{app:sets_classification_prompts}.

\begin{quote}
The Social-Ecological-Technological Systems (SETS) framework analyzes entities across three interconnected dimensions:

Social: human aspects such as community interactions, governance, economic systems, cultural values, and social equity

Ecological: the natural environment and its components, which are often involved in biophysical processes, including natural resources, ecosystem functions, and environmental conditions

Technological: human-made systems and engineered infrastructures, including infrastructure, technological tools, and innovations

The framework can be used to classify entities and concepts based on their alignment with these dimensions. When doing so, it can be helpful to consider not only the entity but the surrounding context in which it was mentioned.

Consider the following context and entity:

Context: The spring was compressed too much
Entity: ``spring''

On a scale from 1-9, where 1 corresponds to the entity having no ecological characteristics and 9 corresponds to extremely high ecological characteristics, given the context, the entity ``spring'' score on the ecological dimension is: 
\end{quote}

We then measure surprisal for $X \in \{1, 2, 3, \ldots, 9\}$ and construct surprisal curves such as those shown in Figure~\ref{fig:sets_classification_bug_insect}.

\subsubsection{SETS Classification Key Findings}

To quantify observations across the full dataset, Table~\ref{tab:02d-aggregate} reports the mean absolute error (MAE) between each model's surprisal-optimal scores and expected scores across all 90 entities in the dataset, broken down by SETS dimension. These results use a standardized set of four Qwen2.5 models. Lower MAE indicates better alignment with expected scores. The results show a clear relationship between model size and accuracy: the 14B variants achieve the lowest MAE (1.43--1.45), followed by 7B-Instruct (1.83), while the 3B-Instruct model performs substantially worse (2.95). The 14B base model slightly outperforms the 14B instruction-tuned variant overall, driven primarily by its strong performance on the ecological dimension (MAE = 0.98). We note that the expected scores are researcher-assigned values rather than multi-rater consensus scores, so these results should be interpreted as alignment with one researcher's expectations rather than an objective benchmark.

\begin{table}[t]
\centering
\caption{SETS classification MAE by model and dimension ($n=90$ entities). Lower is better. All models are Qwen2.5.}
\label{tab:02d-aggregate}
\footnotesize
\begin{tabular}{lrrrr}
\toprule
\textbf{Model} & \textbf{Soc.} & \textbf{Ecol.} & \textbf{Tech.} & \textbf{Mean} \\
\midrule
3B-Instruct & 2.92 & 2.99 & 2.94 & 2.95 \\
7B-Instruct & 2.21 & 1.51 & 1.78 & 1.83 \\
14B-Instruct & 1.86 & 1.17 & 1.33 & 1.45 \\
14B & 2.04 & 0.98 & 1.27 & 1.43 \\
\bottomrule
\end{tabular}
\end{table}

\begin{figure*}[t]
\centering
\includegraphics[width=\textwidth]{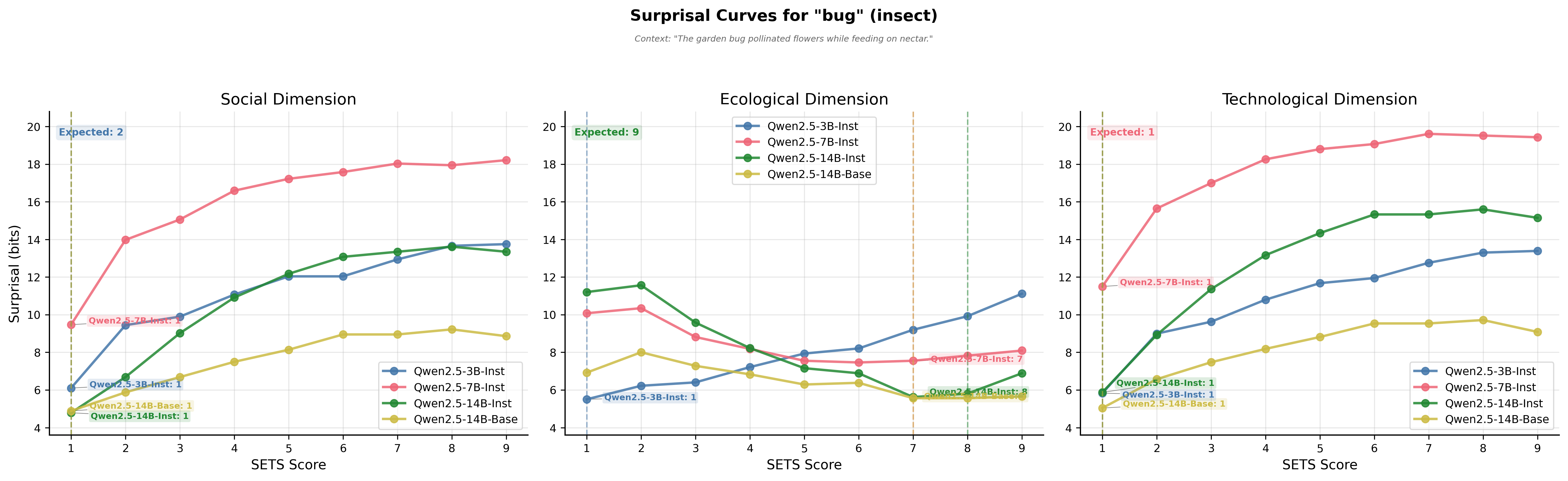}
\caption{Surprisal curves for ``bug'' (insect meaning) on the social, ecological, and technological dimensions with moderate context (``The garden bug pollinated flowers while feeding on nectar''). The 14B models correctly identify high ecological scores.}
\label{fig:sets_classification_bug_insect}
\end{figure*}

\begin{figure*}[t]
\centering
\includegraphics[width=\textwidth]{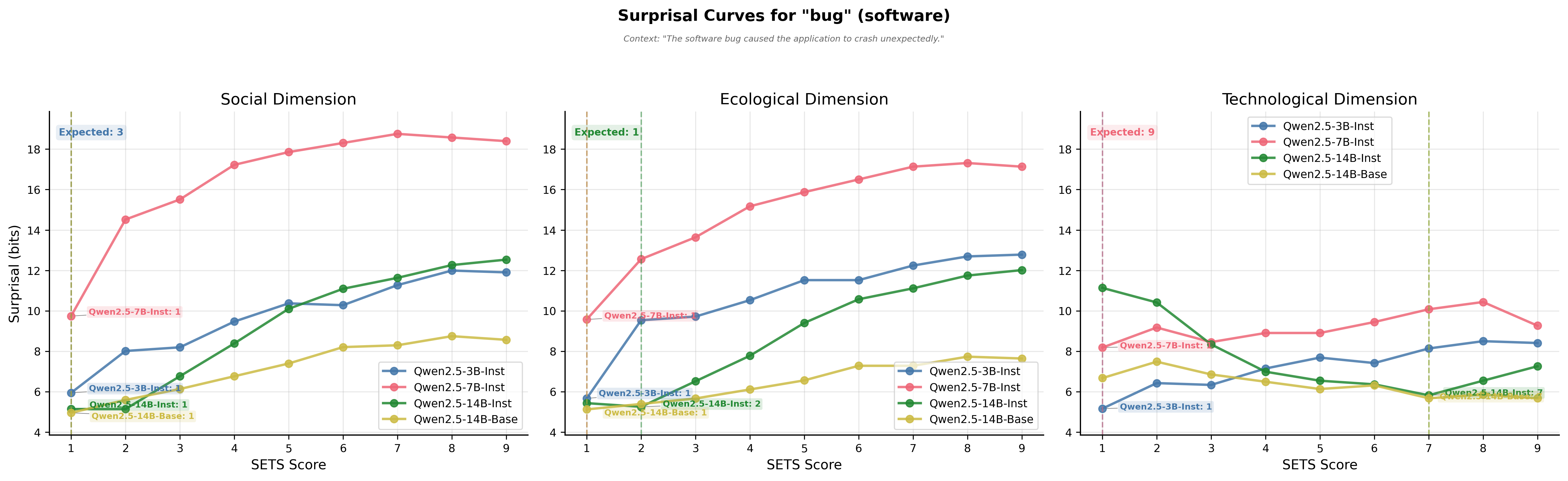}
\caption{Surprisal curves for ``bug'' (software meaning) on the social, ecological, and technological dimensions with moderate context (``The software bug caused the application to crash unexpectedly''). The 14B models correctly identify high technological scores.}
\label{fig:sets_classification_bug_software}
\end{figure*}

Surprisal curves often showed clear minima corresponding to expected scores indicating the LLMs were able to distinguish between homonymous entities. For example, we tested the word ``bug'' and found that the insect meaning shows minimum surprisal at 7--8 on the ecological dimension but 1 on the technological dimension (Figure~\ref{fig:sets_classification_bug_insect}). Yet, as one would hope, when the entity is a software bug, the pattern flips for most models: the minimum surprisal is at 7 on the technological dimension and 1--2 on the ecological dimension (Figure~\ref{fig:sets_classification_bug_software}). The figures show results for four Qwen2.5 models (3B-Instruct, 7B-Instruct, 14B-Instruct, and 14B base). Testing multiple models from the same family helps capture possible performance differences associated with model size. In particular, the 3B model assigns a score of 1 across all dimensions for both meanings of ``bug,'' suggesting it cannot reliably disambiguate the entity at this parameter scale. The 7B model performs better on the ecological dimension for the insect meaning (score 7) but still fails on the technological dimension for the software meaning (score 1). The 14B models correctly identify the distinguishing dimensions for both meanings, suggesting that larger models have more robust representations of entity semantics.

\begin{figure*}[t]
\centering
\includegraphics[width=\textwidth]{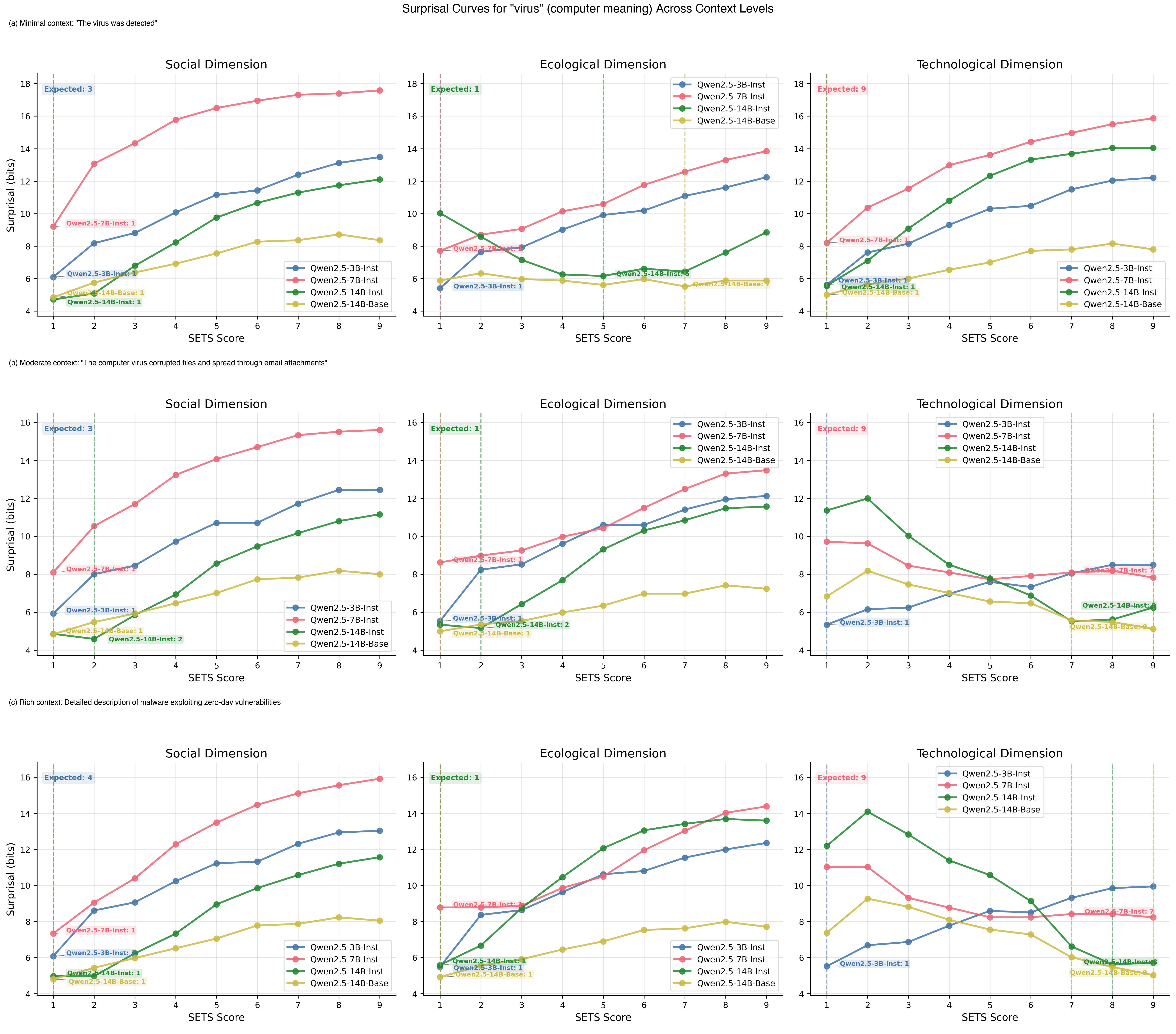}
\caption{Surprisal curves for ``virus'' (computer meaning) across three context levels: (a) minimal context, where models interpret ``virus'' as biological and assign high ecological scores; (b) moderate context, where 14B models correctly shift to high technological scores; (c) rich context describing malware behavior, indicating context-dependent disambiguation. The 3B model never adjusts its technological score regardless of context.}
\label{fig:sets_classification_virus_context}
\end{figure*}

Context of the entity also plays a crucial role in determining the optimal scores. For example, we took three instances for the entity ``virus'' (as in a computer virus) and found that the technological score changed dramatically depending on the context (Figure~\ref{fig:sets_classification_virus_context}). With minimal context (``The virus was detected''), all models assign a technological score of 1, and the 14B models instead assign higher ecological scores (5--7), suggesting they interpret ``virus'' as biological. However, as we add more context making it clearer that we are discussing a computer virus, the pattern flips: with moderate context (``The computer virus corrupted files and spread through email attachments''), the technological score jumps to 9 for the 14B models and 7 for the 7B model, while the ecological score drops to 1--2. The rich context (a detailed description of malware exploiting zero-day vulnerabilities) also displayed this pattern with technological scores of 8--9 for the 14B models. The 3B model never picked up on the technological dimension regardless of context, assigning a score of 1 across all three context levels.

\subsection{Causal Statement Identification}

Causal reasoning is fundamental to human cognition and critical for AI systems \citep{pearl2009causality}. As such, there is general interest in exploring the extent to which LLMs can identify or internally represent causal relationships \citep{kiciman2023causal}. If they can do so, then that may suggest elements of functional world models. To investigate this, we test how well surprisal measurements can provide insights into an LLM's ability to distinguish causal from non-causal statements. We do this by framing the task either as a binary classification task or an ordinal-scaled rating task. 

\subsubsection{Binary Task Description}

For this task, the model must decide whether or not a given statement expresses a causal relationship. Under this binary classification task framing, for each statement, we can measure the extent to which the LLM is surprised by a ``True'' or ``False'' completion to a prompt such as:

\begin{quote}
\texttt{[Causal relationship definition]}\\
\texttt{Statement: ``Smoking causes lung cancer.''}\\
\texttt{This statement expresses a causal relationship:}
\end{quote}

In theory, that sounds simple. In practice, it is more nuanced. For example, how much context should we provide to the model about what a causal relationship is? And what should the response format be? As an initial proof of concept, we test three context levels (full causal background, minimal definition, no context) and multiple response formats (True/False, Yes/No, binary choice). The full context provides a detailed definition of causality and causatives, the minimal context provides a brief definition, and the no context provides no definition. We test the different binary scales to account for potential variations in how the models might interpret the task and options. This is akin to a robustness analysis to check whether the model's performance is sensitive to the specific wording of the choices. It is also the experiment most aligned with the minimal pairs experiment setup. The more complete set of prompts is provided in Appendix \ref{app:causal_classification_prompts}. 

For this set of experiments, we use a synthetic test dataset generated by the researcher to intentionally range across a spectrum from clear-cut to ambiguous cases. The statements can be categorized into five groups, ranging on the degree of causality expressed. The groups are: explicit causal (clear causal markers like ``causes'', ``results in'', etc.), implicit causal (implied causation like ``if-then'' statements), correlational (temporal/statistical association), non-causal (descriptive), and ambiguous (borderline cases). The distribution of those was: 37 causal, 45 non-causal, 18 correlational. For binary classification, we treated explicit and implicit causal as positive and all others as negative, yielding 37 positive and 63 negative items. We use that setup to test hypotheses about model performance under different levels of difficulty with the statements. Examples of implicit causal statements can include ``If it rains, the ground will get wet'' and ``If I study hard, I will get an A in class''. In contrast, associational statements can include sentences such as ``As temperatures rose, the ice cream sales increased.''

\subsubsection{Binary Task Key Findings}

\begin{figure*}[t]
    \centering
    \includegraphics[width=\textwidth]{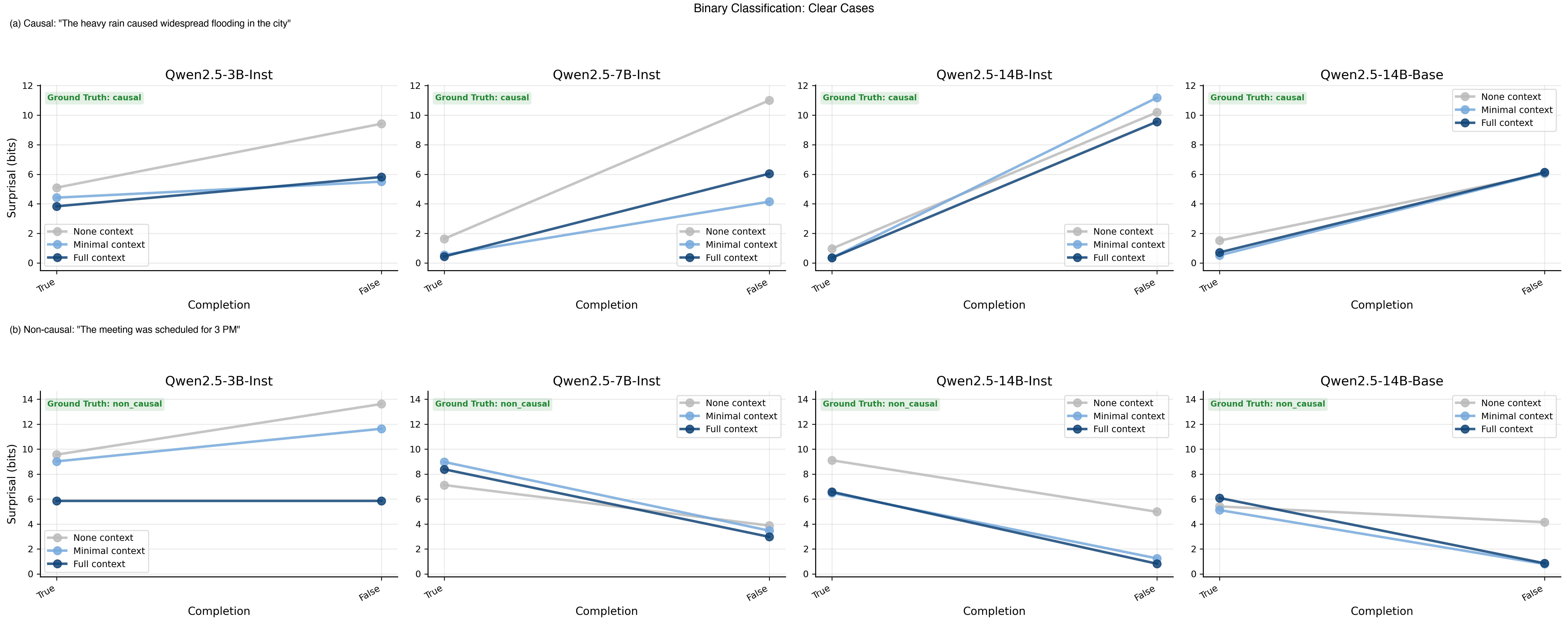}
    \caption{Binary classification surprisal for clear cases: (a) the causal statement ``The heavy rain caused widespread flooding in the city,'' where upward-sloping lines indicate lower surprisal for ``True''; (b) the non-causal statement ``The meeting was scheduled for 3 PM,'' where downward-sloping lines indicate lower surprisal for ``False.'' Each panel shows one model with three context levels.}
    \label{fig:causal-binary-clear}
\end{figure*}

Using the same four Qwen2.5 models as the SETS experiment, we tested these models with three binary options (True/False, Yes/No, Binary Choice) and three context levels (full causal background, minimal definition, no context). Figure~\ref{fig:causal-binary-clear} shows results for two clear-cut cases. For the causal statement ``The heavy rain caused widespread flooding in the city'' (panel a), models demonstrate lower surprisal for the ``True'' completion across all models and context levels, as shown by lines sloping upward from left to right. The non-causal statement ``The meeting was scheduled for 3 PM'' (panel b) shows the opposite: lines slope downward, indicating lower surprisal for ``False.'' This suggests that the models are not simply biased toward one classification by the task framing.

\begin{figure*}[t]
    \centering
    \includegraphics[width=\textwidth]{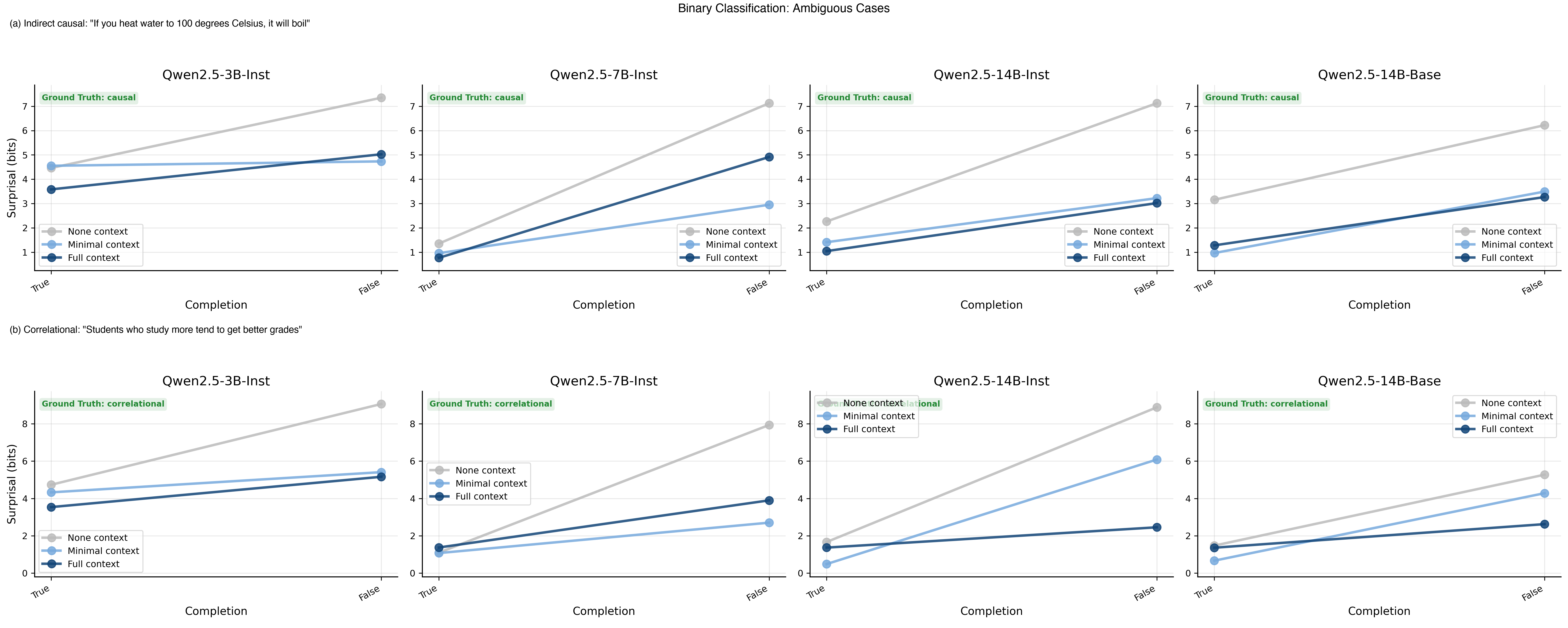}
    \caption{Binary classification surprisal for ambiguous cases: (a) the indirect causal statement ``If you heat water to 100 degrees Celsius, it will boil,'' where lines slope upward but less steeply than clear cases; (b) the correlational statement ``Students who study more tend to get better grades,'' where mixed slopes reflect genuine ambiguity. Compare with Figure~\ref{fig:causal-binary-clear}.}
    \label{fig:causal-binary-ambiguous}
\end{figure*}

More revealing are the ambiguous cases shown in Figure~\ref{fig:causal-binary-ambiguous}. For the indirect causal statement ``If you heat water to 100 degrees Celsius, it will boil'' (panel a), most models and context levels still demonstrate lower surprisal for ``True,'' but the lines are less steep than in the clear causal case, indicating greater model uncertainty. This reduced steepness reflects higher entropy---the probability mass is more spread out across completions rather than concentrated on a single answer. For the correlational statement ``Students who study more tend to get better grades'' (panel b), the pattern is more mixed: while most models still lean toward ``True,'' some context levels produce nearly flat lines. The ground truth label for this statement is ``correlational,'' and the varying slopes reflect genuine ambiguity: the 14B-Instruct model shows relatively consistent upward slopes regardless of context, while the 14B base model shows more variation across context levels.

Table~\ref{tab:02e-aggregate} reports aggregate classification accuracy across the full dataset of 100 statements, including explicit causal, implicit causal, correlational, non-causal, and ambiguous categories. These are broken down by model and context level. The results use a standardized set of four Qwen2.5 models and were consistent with the patterns observed in the individual examples above: larger models achieve higher accuracy, and the effect of context depends on model size. For the 3B-Instruct model, providing full context improves accuracy substantially (68.0\% vs.\ 44.7\% with no context), whereas the 14B-Instruct model shows minimal sensitivity to context level (76.7--78.0\%). The 14B base model shows a noteworthy pattern where the no-context condition (77.0\%) outperforms the minimal-context condition (68.3\%), suggesting that incomplete context definitions may interfere with the base model's prior representations of causality.

\begin{table}[t]
\centering
\caption{Causal binary classification accuracy (\%) by model and context level ($n=100$). Accuracy averaged across response formats. All models are Qwen2.5.}
\label{tab:02e-aggregate}
\small
\begin{tabular}{lrrrr}
\toprule
\textbf{Model} & \textbf{Full} & \textbf{Minimal} & \textbf{None} & \textbf{Mean} \\
\midrule
3B-Instruct & 68.0 & 47.3 & 44.7 & 53.3 \\
7B-Instruct & 75.7 & 72.7 & 69.3 & 72.6 \\
14B-Instruct & 78.0 & 77.0 & 76.7 & 77.2 \\
14B & 76.0 & 68.3 & 77.0 & 73.8 \\
\bottomrule
\end{tabular}
\end{table}

\subsubsection{Ordinal-Scaled Task Description}

Building on binary classification, one might want a more granular expression of uncertainty or recognize that there might be finer nuances in the way that people express causal relationships. To capture that gradation in expression, we tested whether we could extract granular confidence estimates by measuring surprisal across numerical scales. This tests whether surprisal can provide calibrated uncertainty quantification beyond the binary choices tested in the preceding experiment.

\subsubsection{Ordinal-Scaled Task Methodology}

Expanding to an ordinal scale raises practical design questions. For example, how many points should the scale have, even or odd, five or nine, etc. Additionally, what should the anchors be? Should they be equidistant or should they be more spread out? Should the scale be bipolar (i.e., capturing both causal and non-causal content) or unipolar (i.e., capturing only causal content)? For an initial approach into this area, we tested two of the five framings listed in Appendix \ref{app:causal_classification_prompts}: bipolar causality and causal strength. The others were explored in preliminary work but are not reported here. For those two framings, we test on a five-point scale and a nine-point scale. With each scale and each framing, we specify anchor point values. For example, for the causal strength scale, the anchor points are 1 = no causal content and 5 = very strong causal content. For each statement and framing, we measure surprisal for every scale position completion (e.g., ``1'', ``2'', ``3'', ``4'', ``5'') and consequently extract the full surprisal distribution. We perform this measurement for varying levels of background information on causal statements (context levels), specifically testing three levels (no information, minimal information, and full information), as was also the case with the binary classification task. 

For the minimal context variation of the experiment, a representative prompt looks like this:

\begin{quote}
A causal relationship exists when one event, action, or state brings about, influences, or determines another. Causal relationships can be expressed through explicit markers (because, causes, leads to) or implied through conditional statements and purpose expressions.

How strong is the causal content in this statement:
``Monitoring stations indicate that heavy rainfall led to widespread flooding in low-lying areas, according to reports.''

Rate from 1 to 5:
1 = No causal content
5 = Very strong causal content

Rating:
\end{quote}

For each scale length, we measure surprisal for every position completion (1 through 5 or 1 through 9) and repeat across all scale framings and context levels.

\subsubsection{Ordinal-Scaled Task Key Findings}

To enable direct comparison with the binary experiment, we revisit the same statements on ordinal scales. For the causal statement ``The heavy rainfall caused widespread flooding in the city,'' the surprisal curves across all model, context, and scale combinations are predominantly monotonically decreasing---models find higher causal ratings less surprising (Figure~\ref{fig:causal_statement_surprisal_curves_simple_statement} in Appendix~\ref{app:ordinal_curves}). This monotonicity indicates internal consistency: models that preferred a rating of 5 also found 4 less surprising than 3, and so on. Different scale framings (causal strength and bipolar causality) showed convergent patterns for this clear-cut case.

For the correlational statement ``Students who study more tend to get better grades,'' the curves shift to more parabolic shapes with minima falling in the middle of the scales rather than at the extremes (Figure~\ref{fig:causal_statement_surprisal_curves_ambiguous_statement} in Appendix~\ref{app:ordinal_curves}). This suggests that the models find moderate ratings less surprising than those indicating the statement definitely was or was not causal, which is consistent with the genuine ambiguity of a statistical association. The 14B base model showed flatter surprisal curves than the instruction-tuned models for this statement, suggesting greater uncertainty. Exploring how fine-tuning affects model uncertainty on ambiguous items is an area for future work.

Beyond these qualitative patterns, we can also quantify overall performance on this task. Table~\ref{tab:02f-aggregate} reports directional accuracy (the percentage of statements where the surprisal minimum falls on the correct side of the scale midpoint) across the full dataset of 100 statements, using a standardized set of four Qwen2.5 models. Results are broken down by context level and scale framing (CS = causal strength, BC = bipolar causality). Accuracy is generally high across all conditions (74--92\%), suggesting that the ordinal scaling approach preserves the discriminative power observed in binary classification while offering more granular uncertainty information, in general. The causal strength framing tended to slightly outperform the bipolar causality framing, and larger models maintain their advantage, with the 14B-Instruct model achieving the highest mean accuracy (89.0\%).

\begin{table*}[t]
\centering
\caption{Causal scaled directional accuracy (\%) by model, context level, and framing ($n=100$). CS = causal strength, BC = bipolar causality. All models are from the Qwen2.5 family.}
\label{tab:02f-aggregate}
\begin{tabular}{lrrrrrrr}
\toprule
\textbf{Model} & \textbf{Full/CS} & \textbf{Full/BC} & \textbf{Min./CS} & \textbf{Min./BC} & \textbf{None/CS} & \textbf{None/BC} & \textbf{Mean} \\
\midrule
3B-Instruct & 82.0 & 82.0 & 79.0 & 85.0 & 81.0 & 74.0 & 80.5 \\
7B-Instruct & 86.0 & 83.0 & 84.0 & 83.0 & 87.0 & 78.0 & 83.5 \\
14B-Instruct & 90.0 & 91.0 & 91.0 & 88.0 & 88.0 & 86.0 & 89.0 \\
14B & 88.0 & 87.0 & 89.0 & 86.0 & 92.0 & 85.0 & 87.8 \\
\bottomrule
\end{tabular}
\end{table*}

\subsection{Figurative Language Detection}

Figurative language provides a signal about how people conceptualize topics \citep{lakoff1980metaphors}. For instance, describing LLMs as ``stochastic parrots'' \citep{bender2021dangers} versus ``a gold mine of opportunities'' reveals very different mental models. Detecting whether language is figurative or literal is thus a natural task for surprisal-based evaluation and one that directly tests the extended minimal pairs paradigm. We believe this is especially true since figurative and literal statements can share surface-level lexical overlap while differing in meaning.

\subsubsection{Task Description}

We test whether surprisal can distinguish figurative from literal language using a paired minimal-pairs design. Each pair consists of a figurative statement and a literal counterpart that shares key lexical items but uses them literally. For example:

\begin{itemize}[leftmargin=*]
    \item \textbf{Figurative:} ``The words hung in the air between them. Neither wanted to reach out and grab them, afraid of what might happen if they acknowledged the truth.'' (metaphor---words as physical objects)
    \item \textbf{Literal:} ``The banner hung in the air between them, suspended by wires. Neither wanted to reach out and grab it, afraid they might tear the delicate fabric.'' (literal---physical object)
\end{itemize} 

Both statements contain ``hung in the air'' and ``reach out and grab,'' but only the first uses these phrases figuratively. The dataset contains 30 such pairs (60 statements total), spanning metaphor, simile, personification, and analogy across domains including business, psychology, nature, and technology. This is a small proof-of-concept test set that will be built upon in future experiments.

\subsubsection{Methodology}

For each statement, we measure surprisal on a metaphor intensity scale (1 = completely literal, 5 = highly metaphorical) and identify the scale position with minimum surprisal. The primary metric is the \emph{scale discrimination rate}: for each pair, does the figurative statement receive a higher minimum-surprisal position than its literal counterpart? This directly tests whether surprisal measurement can discriminate figurative from literal language within controlled minimal pairs. We tested both 5-point and 9-point scales with the same four Qwen2.5 models and two context levels (minimal and full definitions of figurative language). A representative prompt structure is:

\begin{quote}
\texttt{[Figurative language definitions]}

On a scale from 1 to 5, rate how metaphorical this statement is:

``The words hung in the air between them. Neither wanted to reach out and grab them, afraid of what might happen if they acknowledged the truth.''

1 = Completely literal\\
3 = Somewhat metaphorical\\
5 = Highly metaphorical

Rating:
\end{quote}

The full set of prompts and context levels are described in Appendix~\ref{app:figurative_language_prompts}.

\subsubsection{Key Findings}

\begin{figure*}[t]
    \centering
    \includegraphics[width=\textwidth]{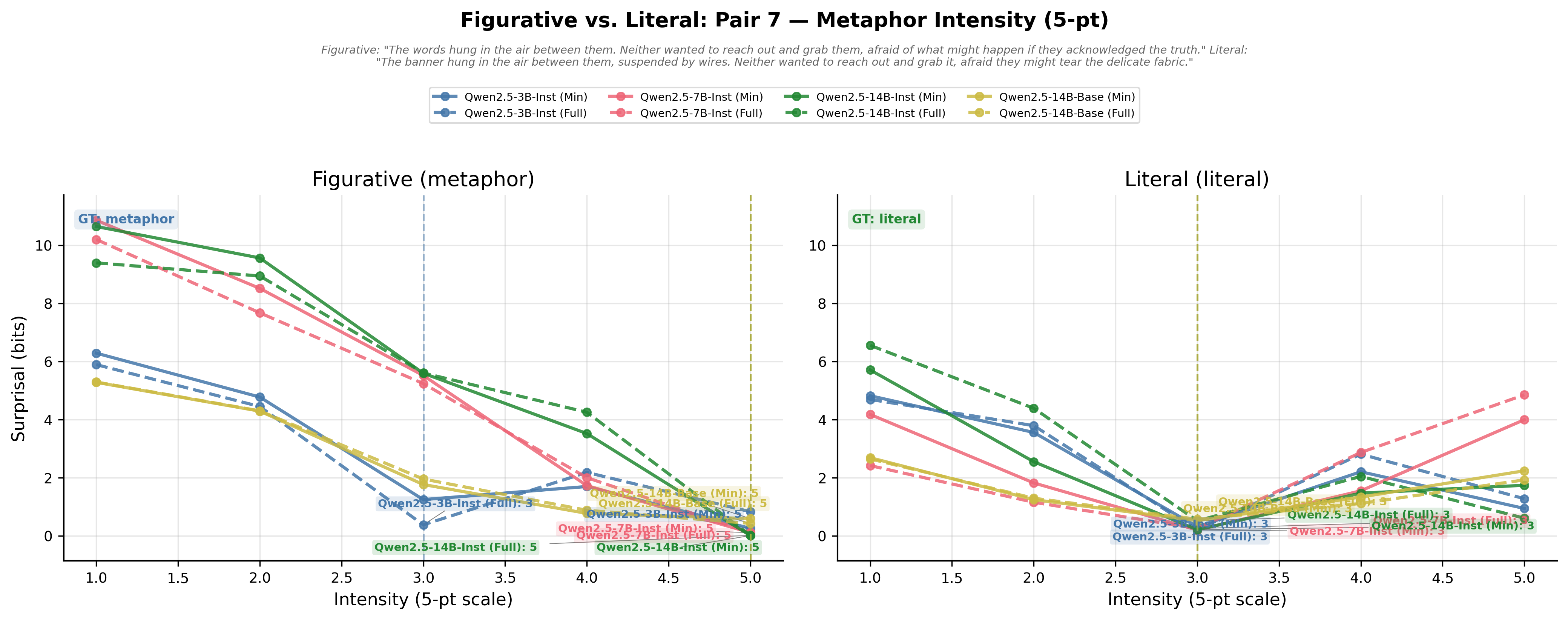}
    \caption{Paired metaphor intensity surprisal curves (5-point scale) for Pair 7. Left: the figurative statement ``The words hung in the air between them...afraid of what might happen if they acknowledged the truth'' shows decreasing surprisal toward high intensity. Right: the literal counterpart ``The banner hung in the air between them...afraid they might tear the delicate fabric'' shows minimum surprisal at low intensity. Both statements share key phrases but differ in figurative status.}
    \label{fig:figurative_paired_scale}
\end{figure*}

Figure~\ref{fig:figurative_paired_scale} illustrates a representative pair on the 5-point metaphor intensity scale. For the figurative statement (left panel), surprisal curves decrease toward the high end of the scale, indicating that models found high metaphor intensity ratings less surprising. For the literal counterpart (right panel), the pattern reversed: curves were relatively flat with minima in the middle of the scale or increasing, with minimum surprisal falling at low or middle intensity scores. This contrast within a single minimal pair, where both statements share the phrases ``hung in the air'' and ``reach out and grab,'' demonstrated that surprisal was sensitive to semantic rather than purely lexical features.

Table~\ref{tab:02g-paired-discrimination} reports the scale discrimination rate across all 30 pairs. On the 5-point scale, the 14B base model achieves 95.0\% mean discrimination, which was the the highest of any model and scale combination. The 7B-Instruct model achieves 90.0\% on the 9-point scale with identical performance across context levels. Discrimination rates are substantially above the 50\% chance baseline across most conditions, suggesting that surprisal-based measurement can reliably distinguish figurative from literal language within controlled minimal pairs.

\begin{table}[t]
\centering
\caption{Figurative language paired scale discrimination rate (\%) by model, scale, and context ($n=30$ pairs). Rate = percentage of pairs where the figurative statement's minimum-surprisal position exceeds the literal counterpart's. All models are Qwen2.5.}
\label{tab:02g-paired-discrimination}
\small
\begin{tabular}{lrrr}
\toprule
\textbf{Model} & \textbf{Minimal} & \textbf{Full} & \textbf{Mean} \\
\midrule
\multicolumn{4}{l}{\textit{5-point scale}} \\
3B-Instruct & 80.0 & 53.3 & 66.7 \\
7B-Instruct & 80.0 & 80.0 & 80.0 \\
14B-Instruct & 70.0 & 63.3 & 66.7 \\
14B & 96.7 & 93.3 & 95.0 \\
\midrule
\multicolumn{4}{l}{\textit{9-point scale}} \\
3B-Instruct & 46.7 & 73.3 & 60.0 \\
7B-Instruct & 90.0 & 90.0 & 90.0 \\
14B-Instruct & 70.0 & 66.7 & 68.3 \\
14B & 76.7 & 63.3 & 70.0 \\
\bottomrule
\end{tabular}
\end{table}

Two patterns in these results are noteworthy. First, the 14B base model's strong performance on the 5-point scale (95.0\%) versus the instruction-tuned 14B model (66.7\%) suggests that instruction tuning may introduce response biases that distort raw surprisal distributions. The base model's surprisal curves may more faithfully reflect the underlying language model's representation of figurativeness. Second, minimal context often outperforms full context on the 5-point scale (e.g., 96.7\% vs.\ 93.3\% for the 14B base model, 80.0\% vs.\ 53.3\% for 3B-Instruct), suggesting that additional definitional context can narrow the surprisal distribution in ways that reduce discriminability.

\subsection{Deductive Coding of Qualitative Survey Responses}

\subsubsection{Task Description}

Qualitative research often involves applying codes to text data \citep{saldana2016coding}. Those codes can be either deductive (i.e., \textit{a priori}, the codes are based on a pre-existing codebook or theoretical framework) or inductive (i.e., the codes are based on the data itself). These codes are labels that researchers would apply to segments of qualitative data to describe patterns (e.g., themes) in those data. The codebook used in our evaluation is shown in Table \ref{tab:codebook-pandemic}. Once a researcher has that codebook, they will typically read the data and make decisions about which code to apply to which segment of data. Thus, one can think of coding as a decision problem where there can be varying degrees of certainty about which code to apply to which segment of data. Here, we test the extent to which surprisal can facilitate deductive coding of survey responses.

\begin{table*}[t]
    \centering
    \small
    \begin{tabular}{|p{4.5cm}|p{10.5cm}|}
        \hline
        \textbf{Code} & \textbf{Definition} \\
        \hline
        Back to normal with no changes/admin & Frustration with administration expecting return to pre-pandemic conditions without implementing improvements \\
        \hline
        Delays, setbacks in careers/research & Career advancement delays, research disruptions, publication setbacks, or professional development impacts \\
        \hline
        Don't want to work in office & Preference to continue remote work rather than returning to office-based work \\
        \hline
        Equity & Concerns about fairness, equal treatment, disparities, or social justice issues related to pandemic recovery \\
        \hline
        Family, personal priorities, childcare & Statements about family responsibilities, childcare arrangements, or balancing personal priorities with work \\
        \hline
        Financial/support & Financial concerns, economic impacts, funding issues, or need for institutional support \\
        \hline
        Lack of common purpose & Absence of shared goals, collective mission, or unified direction in work or institutional settings \\
        \hline
        Not over, less hope of recovering & Pessimism about pandemic recovery, feeling that the crisis is ongoing with little hope for improvement \\
        \hline
        Online teaching challenges & Difficulties with remote teaching, online education delivery, technology issues, or virtual learning environments \\
        \hline
        Public safety (masks, vaccines, etc.) & Statements about public health measures, safety protocols, mask requirements, vaccination policies, and related safety measures \\
        \hline
        Return to campus, students, teaching & Statements about returning to in-person campus activities, student interactions, or classroom teaching \\
        \hline
        Supporting others, employees, students & Responsibilities for supporting colleagues, employees, students, or others during recovery \\
        \hline
        Work/life boundaries & Challenges with maintaining separation between work and personal life, especially during remote work \\
        \hline
    \end{tabular}
    \caption{Codebook for Deductive Coding of Faculty Pandemic Survey Responses (13 codes used in evaluation)}
    \label{tab:codebook-pandemic}
\end{table*}

\subsubsection{Methodology}

Given a codebook with codes and definitions for the codes along with short text to code (i.e., label with the codes, such as open-ended survey responses), we measure surprisal for code applicability scores. We tested both quantitative scales (1-9 scale, where 1 = not applicable, 9 = highly applicable) and qualitative scales. For the qualitative scales, we constrained scale labels to single tokens to avoid multi-token averaging. Qualitative scale labels included degree-of-applicability (e.g., ``not'', ``somewhat'', ``very'', ``extremely''). An example of the prompt and framing used for the deductive coding experiment is shown in Appendix \ref{app:deductive_coding_prompts}. 

As shown in the Appendix \ref{app:deductive_coding_prompts}, experiments for this task can systematically vary the scale options, the number of scale options, the scale type (quantitative or qualitative), background information on what qualitative coding is, persona assignments, and prompt formatting (i.e., use of XML tags or other headings), whether to give the LLM the code and definition or only the code, and the size of the segment of text to code. By altering these variables, we can also test the robustness of surprisal-based coding and capture degrees of uncertainty in the coding process. Hypothetically, that uncertainty could be a useful signal for human coders to review the LLM's coding decisions. 

For the standardized evaluation, we used texts from a survey of faculty members returning to campus toward the end of the pandemic, with a codebook reflecting themes such as work/life balance, career advancement, and public safety. We tested four Qwen2.5 models on a balanced dataset of 40 text--code pairs using a 1--5 code applicability scale (1 = not applicable, 5 = highly applicable), with two persona conditions (no persona and qualitative researcher).

\subsubsection{Key Findings}

\begin{figure*}[t]
    \centering
    \includegraphics[width=\textwidth]{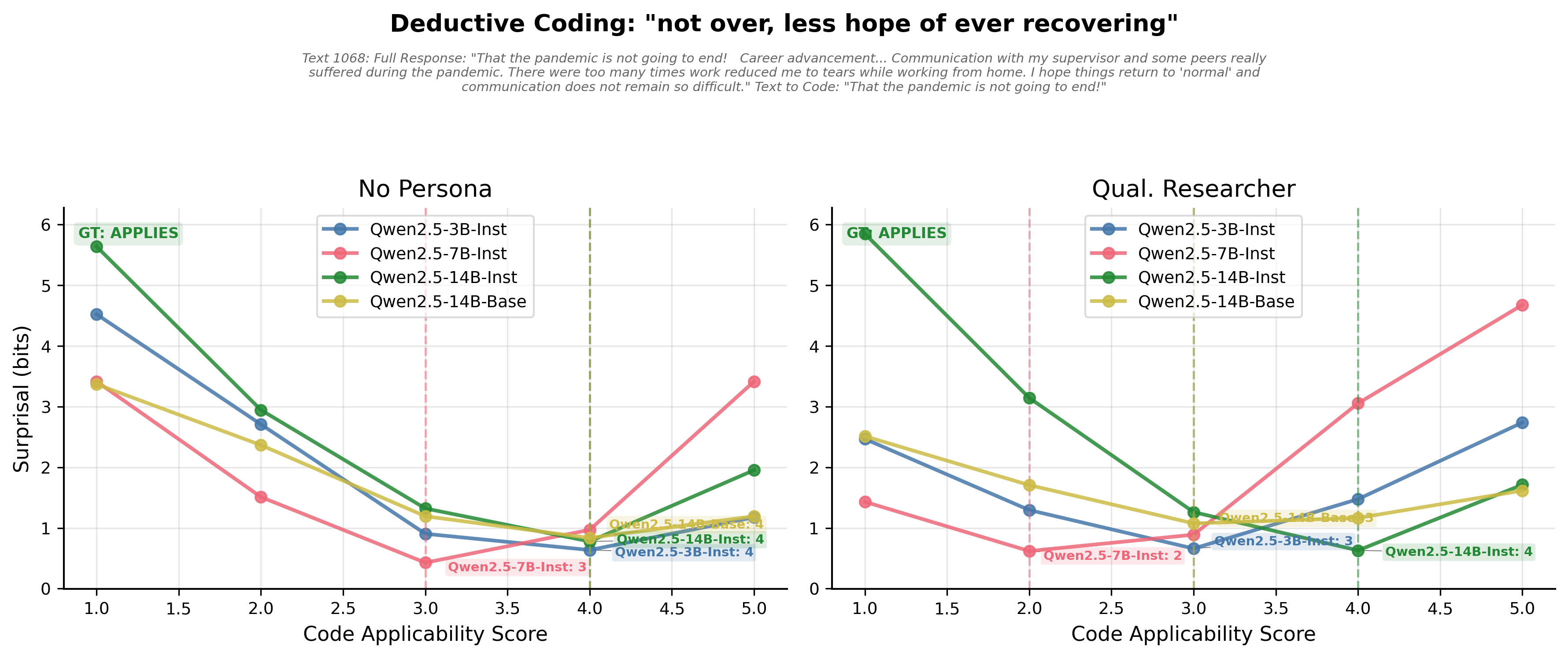}
    \caption{Surprisal-based coding for the code ``not over, less hope of ever recovering'' applied to a text about the pandemic not ending. Models correctly find minimum surprisal at scores 3--4 (applicable).}
    \label{fig:deductive-coding-not-over-positive}
\end{figure*}

Below we present figures with two panels (one per persona condition) with four model curves, allowing direct comparison of how model size and persona assignment affect coding judgments. Figure \ref{fig:deductive-coding-not-over-positive} shows a positive case where the code ``not over, less hope of ever recovering'' was tested against the text ``That the pandemic is not going to end!'' Most models correctly identify this code as applicable, with minimum surprisal falling at scores of 3--4 across most models and persona conditions. The bowl-shaped curves with interior minima indicate that the models found moderate applicability scores less surprising than the extremes of no applicability or high applicability, suggesting belief that the code applies but that it might not be a clear-cut case.

\begin{figure*}[t]
    \centering
    \includegraphics[width=\textwidth]{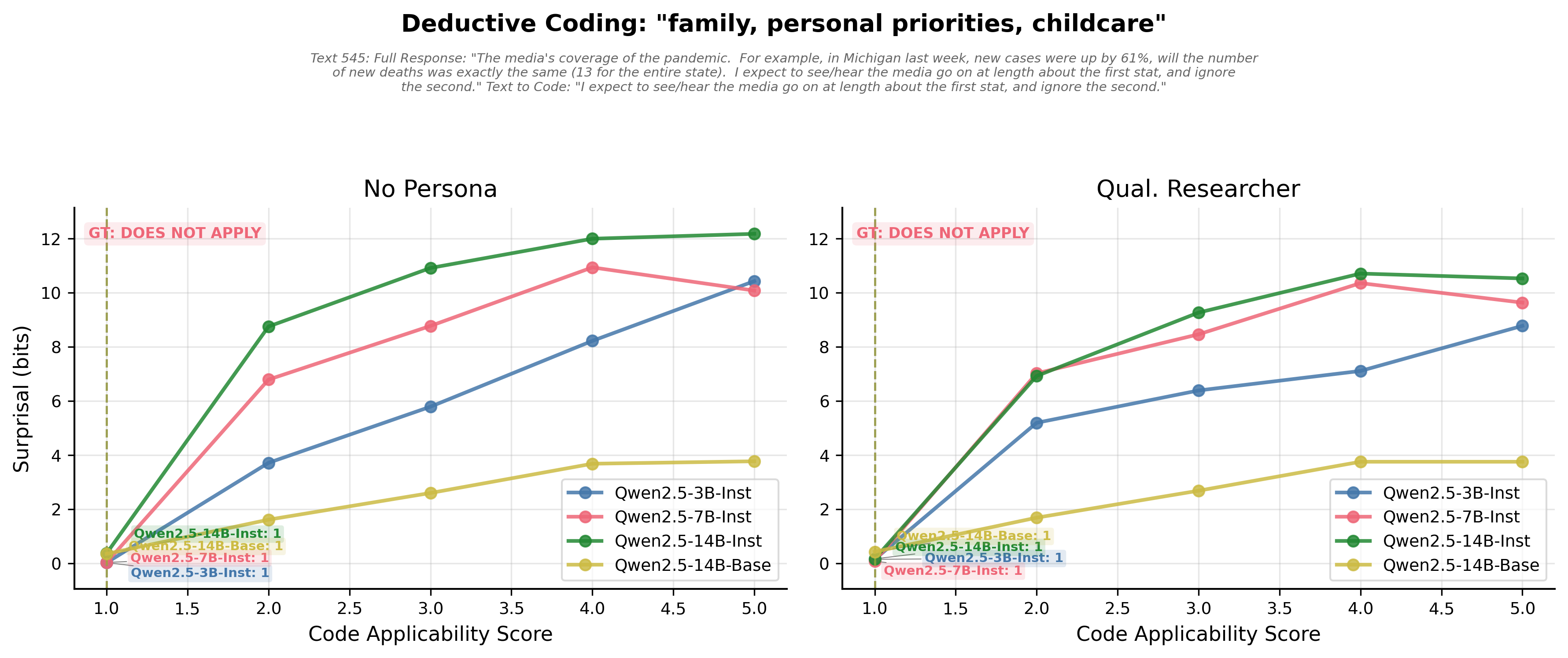}
    \caption{Surprisal-based coding for the code ``family, personal priorities, childcare'' applied to a text about pandemic media coverage. Models correctly find minimum surprisal at score 1 (not applicable).}
    \label{fig:deductive-coding-family-negative}
\end{figure*}

To check whether the models are simply biased toward finding all codes applicable, we can examine a negative case. Figure \ref{fig:deductive-coding-family-negative} shows the code ``family, personal priorities, childcare'' tested against text about pandemic media coverage. Here, we see a reversal of the prior pattern: surprisal increases monotonically from score 1 to 5, with all models finding score 1 (not applicable) as the least surprising option (with the exception of the 7B model for the no persona prompt, where the 4 and 5 surprisals do not follow that monotonic trend). This consistency across models and persona conditions provides a good check that the models are not simply following a fixed pattern of surprisal scores.

\begin{figure*}[t]
    \centering
    \includegraphics[width=\textwidth]{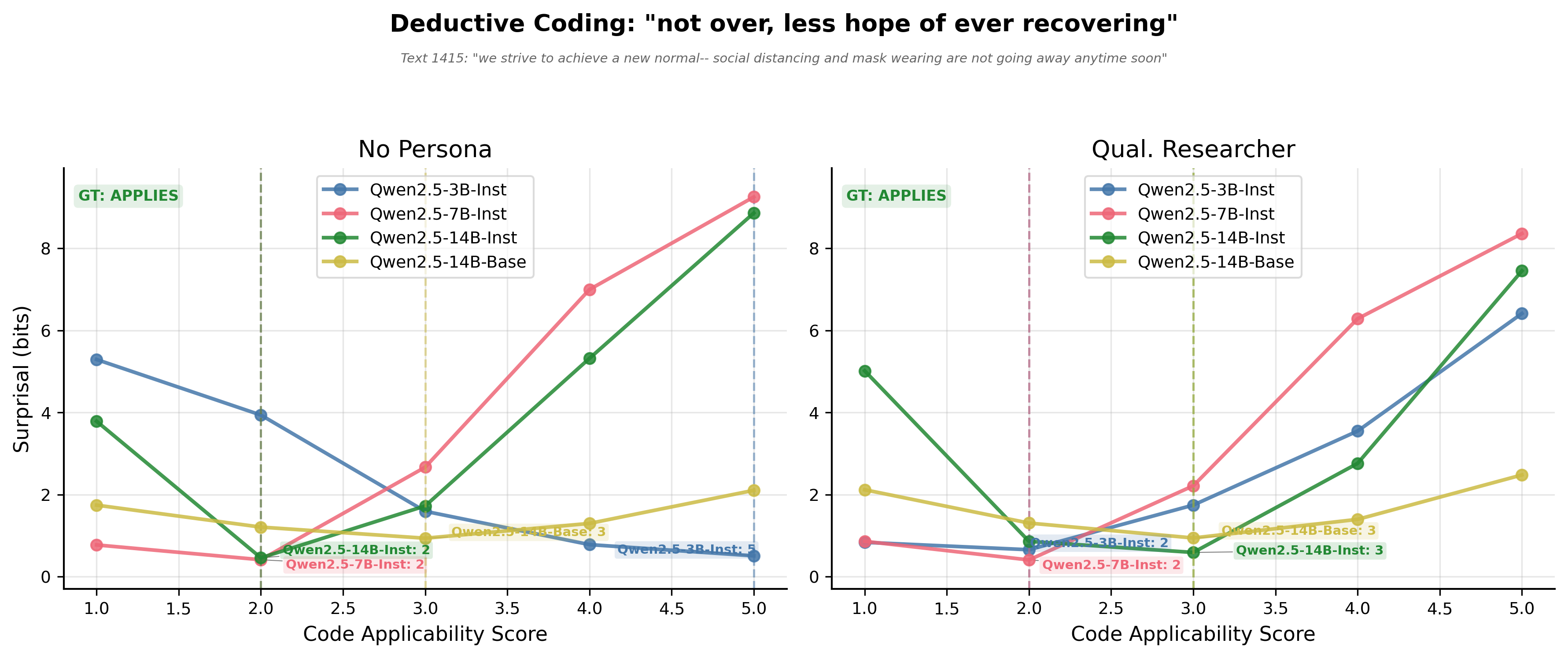}
    \caption{Surprisal-based coding for the code ``not over, less hope of ever recovering'' applied to a text about striving for a new normal. Models show varied patterns, with larger models finding minimum surprisal at moderate applicability scores.}
    \label{fig:deductive-coding-not-over-recovering}
\end{figure*}

Figure \ref{fig:deductive-coding-not-over-recovering} shows another positive case for the code ``not over, less hope of ever recovering'' with a different text (``we strive to achieve a new normal--- social distancing and mask wearing are not going away anytime soon''). Here, the models show more varied patterns across model sizes. The 14B-Instruct and 14B base models find scores of 2--3 as the minimum, while the smaller models show less consistent patterns. This variation highlights how model size interacts with coding difficulty: when the textual markers are less direct, larger models show more nuanced discrimination while smaller models may struggle.

To quantify performance more systematically, Tables~\ref{tab:02j-accuracy-aggregate} and~\ref{tab:02j-f1-aggregate} report aggregate accuracy and F1 scores across a balanced dataset of 40 text--code pairs, using a standardized set of four Qwen2.5 models. Accuracy is computed by thresholding the surprisal-optimal score at $\geq 3$ on the 1--5 scale (i.e., the model considers the code applicable if the minimum-surprisal position is 3 or higher). The 14B-Instruct model achieves the best performance on both metrics (75.0\% accuracy, 72.2\% F1), consistent with the size-related patterns observed in other experiments. The persona manipulation has inconsistent effects across models: it slightly improves the 3B-Instruct model but slightly hurts the 7B-Instruct model. The discrepancy between accuracy and F1 for the 7B-Instruct model (66.3\% accuracy vs.\ 58.4\% F1) suggests an asymmetry in its error patterns, with more false positives than false negatives.

\begin{table}[t]
\centering
\caption{Deductive coding accuracy (\%) by model and persona ($n=40$, threshold $\geq 3$). All models are Qwen2.5.}
\label{tab:02j-accuracy-aggregate}
\small
\begin{tabular}{lrrr}
\toprule
\textbf{Model} & \textbf{No Pers.} & \textbf{Qual.\ Res.} & \textbf{Mean} \\
\midrule
3B-Instruct & 65.0 & 70.0 & 67.5 \\
7B-Instruct & 67.5 & 65.0 & 66.3 \\
14B-Instruct & 75.0 & 75.0 & 75.0 \\
14B & 67.5 & 67.5 & 67.5 \\
\bottomrule
\end{tabular}
\end{table}

\begin{table}[t]
\centering
\caption{Deductive coding F1 score (\%) by model and persona ($n=40$). All models are Qwen2.5.}
\label{tab:02j-f1-aggregate}
\small
\begin{tabular}{lrrr}
\toprule
\textbf{Model} & \textbf{No Pers.} & \textbf{Qual.\ Res.} & \textbf{Mean} \\
\midrule
3B-Instruct & 65.0 & 64.7 & 64.9 \\
7B-Instruct & 60.6 & 56.2 & 58.4 \\
14B-Instruct & 72.2 & 72.2 & 72.2 \\
14B & 66.7 & 64.9 & 65.8 \\
\bottomrule
\end{tabular}
\end{table}

These investigations highlight the promise of surprisal-based coding for deductive coding of survey responses. Future work will explore the robustness of this approach with a particular focus on capturing degrees of uncertainty in the coding process and the extent to which it aligns with human coding decisions. Doing so might have direct applications for qualitative data analysts in addition to providing further indicators of LLM capabilities.

\section{Discussion}
\label{sec:discussion}

Across four domains, three consistent patterns emerged. First, the surprisal-based approach produced interpretable classification signals with clear minima at expected scale positions and monotonic curves across ordinal reasoning in clear-cut cases. Second, entropy over the restricted completion set ($a_i$) tended to flag genuinely ambiguous items, e.g., correlational statements such as ``students who study more tend to get better grades'' or deductive codes with partial textual support tended to have higher entropy while clear-cut items produced peaked distributions with lower entropy. Third, performance generally scaled with model size, though there were notable exceptions: the 14B base model sometimes outperformed its instruction-tuned variant, and smaller models occasionally performed better than the larger models. These findings suggest that the relationship between fine-tuning, parameter count, and surprisal-based accuracy is not strictly monotonic and that tuning or scaling the models may reshape probability distributions in ways that do not always enhance surprisal-based evaluation.

\subsection{Model Behavior Across Domains}

Context sensitivity emerged as a key differentiator across experiments. In the SETS task, providing disambiguating context for homonyms like ``virus'' shifted the 14B models' minimum-surprisal position on the technological dimension from 1 to 9 while the 3B model's scores did not change (Figure~\ref{fig:sets_classification_virus_context}). In the causal binary task, providing full causal definitions improved the 3B model's accuracy from 44.7\% to 68.0\%, but had minimal effect on the 14B-Instruct model (76.7\% to 78.0\%; Table~\ref{tab:02e-aggregate}), and not in the direction one would expect. In the figurative language task, additional definitional context sometimes reduced discrimination rates (e.g., 96.7\% to 93.3\% for the 14B base model on the 5-point scale; Table~\ref{tab:02g-paired-discrimination}). These patterns suggest that context provision is not uniformly beneficial: it helps most when models lack sufficient prior knowledge, but can narrow probability distributions in ways that reduce discriminability when models already represent the distinction well. This aligns with \citet{rauba2024context} on the importance of probing context sensitivity and with \citet{pezeshkpour2024large} on how contextual cues can affect LLM performance in non-obvious ways.

\subsection{Genuine Uncertainty vs. Errors}

A key question for any evaluation method is whether observed uncertainty reflects genuine ambiguity in the task or simply model confusion. Our experiments suggest that entropy-based uncertainty measures may distinguish between these cases. In the causal statement experiments, the statement ``Students who study more tend to get better grades'' produced notably flat surprisal curves across models and contexts, with minimum surprisal values falling in the middle of the scale (see Figure~\ref{fig:causal_statement_surprisal_curves_ambiguous_statement}). This pattern is exactly what one would expect for a genuinely ambiguous case. The statement expresses a statistical association that could plausibly be interpreted as causal or non-causal, and the models' flatter curves reflect that ambiguity.

In contrast, other times, when models were simply wrong (e.g., the 3B model misclassifying ``bug'' in a software context), they showed confident but incorrect responses with low entropy. This asymmetry is encouraging: it suggests that high entropy genuinely signals task difficulty or ambiguity, while low entropy indicates model confidence, though not necessarily correctness. The aggregate accuracy tables suggest that larger models may be better calibrated in this regard. The 14B-instruct model achieves the highest accuracy across all four domains on average while also producing flatter curves for ambiguous cases. For practitioners, this may imply that entropy measures can serve as useful flags for cases requiring human review in a human-in-the-loop setting \citep{wu2022survey}, complementing the primary classification decision. Future work can test for correlations between human judgment of the difficulty for each item and the model's entropy for that item.

\subsection{Framework Design Considerations}

\subsubsection{Binary vs. Ordinal Framings}

Our experiments with causal statement identification tested both binary (True/False, Yes/No) and ordinal (1-5, 1-9) framings of the same underlying task. These different framings may measure different aspects of model judgment. Binary framings force a categorical decision and reveal the model's threshold for classification. Although the surprisal difference between the two options provides a crude confidence measure, the information is limited because a single delta simply tells the researcher about direction and magnitude of preference but nothing about the shape of model uncertainty overall.

Ordinal framings yield richer information through the shape of the entire surprisal curve. A monotonically decreasing curve (as in Figure~\ref{fig:causal_statement_surprisal_curves_simple_statement}) indicates confident classification with internally consistent ordinal reasoning. A bowl-shaped or parabolic curve with a minimum somewhere in the middle (as in some of the plots in the middle column in Figure~\ref{fig:causal_statement_surprisal_curves_ambiguous_statement}) reveals genuine ambiguity or moderate confidence. The steepness of the curve provides additional confidence information: steep curves suggest high certainty, while flat curves suggest uncertainty regardless of where the minimum falls. This finding suggests that practitioners should choose framings based on their information needs. Binary framings suffice for simple yes/no classification tasks. Ordinal framings are preferable when uncertainty quantification is important, when the underlying construct admits degrees rather than categorical distinctions, or when the uncertainty can be used in downstream tasks as an important signal.

\subsubsection{The Tokenization Challenge}

Surprisal-based evaluation is inherently sensitive to tokenization. Because we measure the probability of specific tokens, the choice of completion format directly affects results. For example, `` True'' (with a leading space) and ``True'' (without) are different tokens with potentially different probabilities. In our experiments, we consistently used leading spaces to match natural continuation patterns, but this choice requires explicit documentation. Beyond formatting conversations, we observed potential biases in certain completion formats. In preliminary deductive coding experiments with qualitative scales, the Yes/No scale showed unexpected behavior where ``No'' was consistently selected as least surprising even when the code appeared applicable. This suggests that models may have prior biases toward certain tokens independent of the classification context; this is a phenomenon related to the surface form competition described by \citet{holtzman2021surface}.

These findings underscore the importance of testing multiple response formats as a robustness check and being cautious about interpreting results from any single format. They also motivate the use of numeric scales (1-5, 1-9), though it is unclear whether numeric scales would be less subject to strong prior biases than semantic labels. Moreover, numeric scales do not escape tokenization challenges entirely. Different tokenizers handle multi-digit numbers differently, so a 1--10 or 1--100 scale introduces multi-token completions that complicate clean measurement.

\subsection{Theoretical Implications}

\subsubsection{What Does ``Model Belief'' Mean?}

Throughout this paper, we have used language suggesting that surprisal reveals what models ``believe'' or find ``natural.'' This framing warrants caution. Strictly speaking, surprisal measures the probability assigned to tokens given context and nothing more. Whether this constitutes ``belief'' in any meaningful sense is philosophically contentious \citep{schwitzgebel2011belief,goldman1979justified}.

We can say more confidently that surprisal patterns correlate with human processing difficulty in psycholinguistic research \citep{hale2001probabilistic, levy2008expectation}. To the extent that LLMs trained on human-generated text acquire similar statistical regularities, their surprisal patterns may loosely track what humans find expected or unexpected. However, this is a correlational claim, not an assertion about genuine understanding or belief. For practical purposes, we recommend interpreting surprisal-based results as revealing learned statistical associations rather than making strong claims about model cognition. For now, the framework's value lies in its efficiency and uncertainty quantification, not in resolving deep questions about machine understanding.

\citet{turpin2023language} demonstrate that chain-of-thought explanations can be disconnected from actual model reasoning, implying that explicit prompted answers may not reflect underlying representations. A final speculative conjecture is that surprisal-based measures might provide more direct access to implicit model beliefs, though this hypothesis requires targeted investigation. We speculate that the distinction between surprisal-based and prompting-based evaluation may be analogous in some ways to dual-process theories of cognition \citep{evans2003two}. In these theories, System 1 processing is automatic and intuitive, while System 2 processing is deliberate and analytical. Surprisal-based evaluation may access something analogous to System 1 responses: immediate, reflexive probability assignments based on learned associations. Prompting with explicit reasoning (e.g., chain-of-thought) may analogously engage System 2-like processing, allowing models to deliberate and potentially override initial impressions, for better or worse.

\subsection{Future Directions}
\label{sec:future-directions}

The four experiments presented above demonstrate the core methodology and illustrate design principles, but they also point to multiple directions for future work, as we have been noting throughout the paper.

\subsubsection{Factual Knowledge Evaluation}

A natural extension of the framework is to factual knowledge assessment. Prior work has shown that language models encode substantial world knowledge in their weights \citep{petroni2019language}, but systematic evaluation of that knowledge remains challenging. Surprisal-based evaluation could address this by measuring surprisal for correct versus incorrect completions of factual statements, e.g., comparing surprisal for ``Paris,'' ``London,'' and ``Berlin'' given the prompt ``The capital of France is.'' Lower surprisal for the correct answer would indicate that the model has encoded the relevant factual association. Applied at scale, this could enable systematic mapping of factual knowledge across domains and model versions.

\subsubsection{Bias and Fairness Analysis}

Bias detection is another promising application. Research has documented that language models can encode and amplify societal biases present in training data \citep{bolukbasi2016man}, yet detecting these biases often requires generating text and subjectively evaluating it. Surprisal-based analysis offers a more direct and quantitative alternative by comparing surprisal patterns across demographic groups. For instance, measuring surprisal for gendered pronouns across occupations (e.g., ``The [doctor/nurse] walked into the room. [He/She]'') would reveal whether a model finds certain gender--occupation pairings more ``expected'' than others, directly quantifying stereotypical associations encoded in the model's probability distribution. Because this approach accesses implicit associations rather than explicit generated statements, it may surface subtle biases that are not apparent in generated text.

\section{Limitations}
While our experiments suggest the promise of surprisal-based evaluation, several limitations constrain its current applicability and interpretation on top of the prior ones noted above such as the single model family and synthetic datasets used in some of the experiments.

\subsection{Measurement Constraints}

The current framework imposes several practical constraints. First, completions are limited to single tokens for clean measurement, which restricts response formats. Multi-token answers like ``Strongly agree'' cannot be directly compared with single-token alternatives without averaging or modification. For example, rather than using a scale from 1-100, we are using a scale from 1-9 to avoid potential multi-token completions associated with a two- or three-digit number. While averaging across tokens may be a valid approach, as it is used when studying surprisal of a sentence \citep{rezaii2023measuring,huber2024surprisal}, it is unclear how well that technique would work with these scale-based experiments. Second, scale design involves non-trivial choices: odd versus even scales affect central tendency, anchor wording affects interpretation \citep{chen2015optimal}, and the optimal number of scale points (3, 5, 7, 9) likely varies by task. We tested 5-point and 9-point scales but did not systematically compare their effectiveness. Third, tokenization sensitivity means that results depend on formatting choices (leading spaces, capitalization) that may not be obvious to practitioners. Likewise, what a 1 and 9 correspond to (as dictated by the prompt and anchors) may also be ambiguous or arbitrary. 

\subsection{Calibration and Interpretability}

A key limitation is that model confidence (hypothetically measured by entropy) does not guarantee accuracy. In our experiments, smaller models sometimes showed high confidence (low entropy) on incorrect answers (e.g., the 3B model on SETS scoring tasks). This means low-entropy responses cannot automatically be trusted without validation. Conversely, high-entropy cases signal uncertainty but do not distinguish between genuine task ambiguity and model confusion. Language models are not necessarily calibrated to generate accurate self-reflection on their own confidence estimates \citep{geng2024survey,kadavath2022language}.

We proposed entropy as a proxy for confidence, but it remains an open question whether this relationship generally holds. At a minimum, one would expect the calibration relationship between entropy and accuracy to vary by model, domain, and task type. Establishing reliable calibration would require extensive validation studies, potentially undermining the efficiency gains that motivate the approach. For high-stakes applications, we currently recommend treating surprisal-based results as preliminary assessments requiring human verification rather than final decisions.

Another limitation of this surprisal-based evaluation is the broader interpretability of the approach. Ultimately, surprisal is just a function of the probability distribution over tokens. It does not explain \emph{why} the model prefers certain completions. Moreover, the model might respond differently if asked to generate reasoning, as in a test-time compute setup \citep{snell2024scaling}. Insofar as reasoning traces may provide insights into the model's reasoning process, surprisal-based evaluation does not provide access to this information. However, some evidence suggests those reasoning traces are only loosely coupled to the model's actual decision process or internal representations \citep{arcuschin2025chain}. Additionally, there are inherent biases in which token the models will prefer \textit{a priori} without conditioning on the context, so some of the observed surprisal scores could be due to these biases rather than the context.

\subsection{Access to Token-Level Probabilities}

Surprisal-based evaluation requires access to token-level log-probabilities (logits), which are available when running open-weight models locally (e.g., via HuggingFace Transformers) but may not be exposed by proprietary API-only models. As of this writing, some commercial APIs (e.g., OpenAI) provide limited log-probability access for top-$k$ tokens, while others provide no access at all. This means the framework in its current form is most directly applicable to open-weight models. However, to the extent that API providers increasingly expose log-probabilities (or that open-weight models continue to close the performance gap with proprietary models) this limitation may diminish over time. Researchers who require evaluation of closed models may need to rely on prompting-based methods or approximate probability estimates through repeated sampling, which forfeits the efficiency advantages of single-pass surprisal measurement.

\subsection{Absence of Direct Comparison with Prompting-Based Evaluation}

A notable gap in the present work is that we do not directly compare surprisal-based classification with standard prompting-based classification on the same tasks. While we argue that the two approaches are complementary, this claim would be substantially strengthened by empirical evidence showing (a) whether they agree or diverge, (b) in which conditions each approach excels, and (c) whether their combination yields better performance than either alone. Such a comparison is a priority for ongoing work and would involve running both zero-shot prompting and surprisal-based classification on the same statement sets with the same models, comparing accuracy, agreement rates, and confidence estimates across methods.

\section{Conclusion}
\label{sec:conclusion}

The minimal pairs paradigm has gained adoption as an approach valuable for probing linguistic knowledge in language models, but in most cases its application has largely stopped at binary grammaticality contrasts. This paper argues for extending the paradigm to ordinal scales and for tasks beyond grammaticality predictions. Extending to ordinal scales enables measurements not only of the preferred response but the full surprisal distribution, providing richer evaluation signals suitable for applied classification tasks across domains. Our experiments, which were motivated by ongoing research in mental models such as entity scoring according to the SETS framework, causal reasoning, figurative language detection, and deductive coding, supported this claim. Surprisal curves tended to produce internally consistent classification signals, and calculating the entropy over the token completion set promisingly distinguished genuinely ambiguous items from confident errors.

To be sure, these results are preliminary, and key gaps remain. Most notably, there was an absence of direct comparison with prompting-based methods and there is room for systematic calibration studies. Nonetheless, the framework's core properties of single-pass efficiency, principled uncertainty quantification, and access to implicit model judgments rather than generated rationalizations position it as a complement to existing evaluation approaches. We hope the demonstrations in this paper serve as a starting point for broader model families and domains.

\iftaclpubformat
\section*{Acknowledgments}

This work was supported in part by a grant from the Virginia Tech Academy of Data Science Discovery Fund and NSF grants EEC 2107008, DUE 2300977, and 2339702.
\fi

\bibliography{references}
\bibliographystyle{acl_natbib}

% Appendices: Prompt Templates
\appendix

\section{SETS Classification Prompts}
\label{app:sets_classification_prompts}

\subsection{SETS Framework Context}

The Social-Ecological-Technological Systems (SETS) framework analyzes entities across three interconnected dimensions:

\begin{itemize}[leftmargin=*]
    \item \textbf{Social}: Human aspects such as community interactions, governance, economic systems, cultural values, and social equity
    \item \textbf{Ecological}: The natural environment and its components, often involved in biophysical processes, including natural resources, ecosystem functions, and environmental conditions
    \item \textbf{Technological}: Human-made systems and engineered infrastructures, including infrastructure, technological tools, and innovations
\end{itemize}

The framework can be used to classify entities and concepts based on their alignment with these dimensions. When doing so, it is helpful to consider not only the entity but the surrounding context in which it was mentioned.

\subsection{Prompt Template Structure}

For each entity and dimension, we measure surprisal for scores 1-9 using the following template:

\begin{quote}
\texttt{[SETS Framework Context]}

Consider the following context and entity:

Context: ``[context\_sentence]''

Entity: ``[entity]''

On a scale from 1-9, where 1 corresponds to the entity having no [dimension] characteristics and 9 corresponds to extremely high [dimension] characteristics, the entity ``[entity]'' score on the [dimension] dimension is:
\end{quote}

\subsection{Example Prompts}

\subsubsection{Example 1: Park (Ecological Dimension)}

\begin{quote}
\small
\ttfamily
The Social-Ecological-Technological Systems (SETS) framework analyzes entities across three interconnected dimensions: [...]

Consider the following context and entity:

Context: ``We went to the neighborhood park.''

Entity: ``park''

On a scale from 1-9, where 1 corresponds to the entity having no ecological characteristics and 9 corresponds to extremely high ecological characteristics, the entity ``park'' score on the ecological dimension is:
\end{quote}

We then measure surprisal for completions \texttt{`` 0''}, \texttt{`` 1''}, ..., \texttt{`` 9''}.

\subsubsection{Example 2: Virus (Technological Dimension)}

For the homonym ``virus,'' context determines scoring:

\textbf{Computer virus context:}
\begin{quote}
\small
\ttfamily
Context: ``The computer virus corrupted files and spread through email attachments.''

Entity: ``virus''

On a scale from 1-9, [...] the entity ``virus'' score on the technological dimension is:
\end{quote}

\textbf{Biological virus context:}
\begin{quote}
\small
\ttfamily
Context: ``The virus was detected.''

Entity: ``virus''

On a scale from 1-9, [...] the entity ``virus'' score on the ecological dimension is:
\end{quote}

\section{Causal Statement Classification Prompts}
\label{app:causal_classification_prompts}

\subsection{Context Levels}

\subsubsection{Minimal Context}
\begin{quote}
\small
\texttt{Causality refers to relationships where one event causes or influences another. Look for statements that express how one thing brings about, leads to, or is responsible for another thing.}
\end{quote}

\subsubsection{Full Context}
The full context provides detailed guidance on causal linguistic markers. An abbreviated version is shown below; the complete version includes 18 categories (a--r) covering explicit markers, causal verbs, noun phrases, adverbial clauses, implied causation, causal reasoning, causative types, causal chain complexity, temporal ordering, counterfactual causality, causal strength, discourse-level causality, implicit causality, correlation vs. causation distinctions, multiple causation, causal chains, parallel causation, and branching causation.

\begin{quote}
\small
Causal relationships can be indicated through various linguistic structures and at different levels of language. Be aware of the following indicators:

a) Explicit markers: because, since, therefore, thus, hence, consequently, so that, in order to
b) Causal verbs: cause, result, produce, generate, lead, induce, trigger, prompt
c) Noun phrases: causes, reasons, effects, consequences, outcomes
d) Adverbial clauses: due to, owing to, as a result of, because of
e) Implied causation: If-then constructions, resultative constructions
f) Causal reasoning: expressions of purpose (to, in order to), statements of intention (intend to, aim to)
g) Types of causatives: Lexical, Morphological, Periphrastic
h) Causal chain complexity: Simple (A causes B) or complex (A causes B, which causes C)
i) Temporal ordering: Cause typically precedes effect, but language allows various orderings
j) Counterfactual causality: ``If X hadn't happened, Y wouldn't have occurred''
k) Causal strength and probability: Language expressing degrees of causal influence
l) Discourse-level causality: Causal relationships spanning across sentences
m) Implicit causality: Verbs carrying implicit causal information
n) Correlation vs. Causation: Be particularly careful to distinguish between:
- True causal relationships where one event directly influences another
- Mere temporal correlation or co-occurrence
- Statistical association without clear causation
- Sequential events without proven causation
o) Multiple causation: When multiple distinct causal relationships exist in the same statement
p) Causal chains: When A causes B, which in turn causes C
q) Parallel causation: When multiple causes lead to the same effect
r) Branching causation: When one cause leads to multiple effects
\end{quote}

\subsection{Scale Framings}

We test five different scale framings to explore how question framing affects surprisal patterns:

\subsubsection{Framing 1: Bipolar Causality Scale}
\begin{quote}
\small
\ttfamily
[Context]

On a scale from 1 to 5, rate the causal content of this statement:

``[statement]''

1 = Definitely non-causal
3 = Neutral/uncertain
5 = Definitely causal

Rating:
\end{quote}

\subsubsection{Framing 2: Belief Strength Scale}
\begin{quote}
\small
\ttfamily
[Context]

The following statement expresses a causal relationship: True or False

``[statement]''

On a scale from 1 to 5, how strongly do you believe this answer:

1 = Definitely False (not causal)
5 = Definitely True (causal)

Rating:
\end{quote}

\subsubsection{Framing 3: Probability Scale}
\begin{quote}
\small
\ttfamily
[Context]

What is the probability that this statement expresses causality:

``[statement]''

Rate from 1 to 5:

1 = 20\% probability (very unlikely to be causal)
5 = 100\% probability (very likely to be causal)

Rating:
\end{quote}

\subsubsection{Framing 4: Causal Strength Scale}
\begin{quote}
\small
\ttfamily
[Context]

How strong is the causal content in this statement:

``[statement]''

Rate from 1 to 5:

1 = No causal content
5 = Very strong causal content

Rating:
\end{quote}

\subsubsection{Framing 5: Dual Classification}
\begin{quote}
\small
\ttfamily
[Context]

Rate this statement on both dimensions:

``[statement]''

A) How causal is this statement? (1 to 5)
1 = Not causal at all, 5 = Highly causal

B) How non-causal is this statement? (1 to 5)
1 = Not non-causal at all, 5 = Highly non-causal

A) Rating:
\end{quote}

For the dual classification framing, we measure surprisal for both A and B completions (e.g., \texttt{``A\_1''}, \texttt{``A\_2''}, ..., \texttt{``B\_1''}, \texttt{``B\_2''}, ...).

\subsection{Example with Minimal Context}
\begin{quote}
\small
\ttfamily
Causality refers to relationships where one event causes or influences another. Look for statements that express how one thing brings about, leads to, or is responsible for another thing.

How strong is the causal content in this statement:

``Monitoring stations indicate that heavy rainfall led to widespread flooding in low-lying areas, according to reports.''

Rate from 1 to 5:

1 = No causal content
5 = Very strong causal content

Rating:
\end{quote}

\section{Figurative Language Detection Prompts}
\label{app:figurative_language_prompts}

\subsection{Context Levels}

\subsubsection{Full Context}
\begin{quote}
\small
Figurative language uses words or expressions with meanings different from their literal interpretation to create vivid imagery, comparisons, or emphasis. Key types include:

\begin{enumerate}
\item METAPHORS: Direct comparisons that state one thing IS another thing (e.g., ``Love is a battlefield'')
\item ANALOGIES: Extended comparisons that explain one concept by comparing it to another (e.g., ``The economy works like a machine'')
\item SIMILES: Comparisons using ``like'' or ``as'' (e.g., ``Bright as the sun'')
\item PERSONIFICATION: Giving human characteristics to non-human things (e.g., ``The ocean roared with fury'')
\end{enumerate}

Metaphors create implicit comparisons without using comparison words, while analogies typically involve more detailed structural comparisons between different domains. Similes make explicit comparisons using comparison words, and personification attributes human qualities to non-human entities.
\end{quote}

\subsubsection{Minimal Context}
\begin{quote}
\small
Figurative language uses non-literal meanings to create comparisons or emphasis. Key types:
\begin{itemize}[leftmargin=*]
\item Metaphors: Direct comparisons (X is Y) without ``like/as''
\item Analogies: Extended comparisons explaining concepts through structural similarities
\item Similes: Explicit comparisons using ``like'' or ``as''
\item Personification: Giving human qualities to non-human things
\end{itemize}
\end{quote}

\subsection{Binary Classification Prompts}

\subsubsection{True/False Format - Metaphor Detection}
\begin{quote}
\small
\ttfamily
[Context]

The following statement contains a metaphor: True or False

``Time is money.''

Answer:
\end{quote} 

Completion targets: \texttt{`` True''}, \texttt{`` False''}

\subsubsection{Yes/No Format - Analogy Detection}
\begin{quote}
\small
\ttfamily
[Context]

Does the following statement contain an analogy?

``The brain works like a computer.''

Answer (Yes or No):
\end{quote}

Completion targets: \texttt{`` Yes''}, \texttt{`` No''}

\subsection{Intensity Scale Prompts}

\subsubsection{Metaphor Intensity (1-5 Scale)}
\begin{quote}
\small
\ttfamily
[Context]

On a scale from 1 to 5, rate how metaphorical this statement is:

``Time is money.''

1 = Completely literal
3 = Somewhat metaphorical
5 = Highly metaphorical

Rating:
\end{quote}

Completion targets: \texttt{`` 1''}, \texttt{`` 2''}, \texttt{`` 3''}, \texttt{`` 4''}, \texttt{`` 5''}

\subsubsection{Analogy Strength (1-9 Scale)}
\begin{quote}
\small
\ttfamily
[Context]

On a scale from 1 to 9, rate the analogical content of this statement:

``The economy works like a machine.''

1 = No analogy present
5 = Moderate analogical content
9 = Strong analogy

Rating:
\end{quote}

Completion targets: \texttt{`` 1''} through \texttt{`` 9''}

\subsection{Multi-Category Classification}
\begin{quote}
\small
\ttfamily
[Context]

Does this statement contain a metaphor, analogy, simile, personification, or none of these?

``The wind whispered through the trees.''

Answer:
\end{quote}

Completion targets: \texttt{`` metaphor''}, \texttt{`` analogy''}, \texttt{`` simile''}, \texttt{`` personification''}, \texttt{`` none''}

\section{Deductive Coding Prompts}
\label{app:deductive_coding_prompts}

\subsection{Task Structure}

Deductive coding applies predefined codes (with definitions) to unstructured text. Our prompts systematically vary experimental factors to assess their impact on coding accuracy.

\subsection{Prompt Components}

A deductive coding prompt consists of:

\begin{enumerate}[leftmargin=*]
\item \textbf{Optional Context}: Background on deductive coding methodology
\item \textbf{Optional Persona}: Role assignment (qualitative researcher, domain expert)
\item \textbf{Text Section}: Survey response or text to be coded
\item \textbf{Code Section}: Code name and/or definition
\item \textbf{Task Section}: Instructions and scale for rating
\end{enumerate}

\subsection{Section Delimiter Styles}

\subsubsection{XML-Style Delimiters}
\begin{quote}
\small
\ttfamily
\textless TEXT\_TO\_CODE\textgreater

Survey Response: ``Work/life balance continues to be the biggest challenge, because the amount of work is still elevated and I don't see that changing.''

\textless/TEXT\_TO\_CODE\textgreater

\textless CODE\_INFORMATION\textgreater

Code: work/life boundaries

Definition: Challenges with maintaining separation between work and personal life, especially during remote work

\textless/CODE\_INFORMATION\textgreater
\end{quote}

\subsubsection{ALL CAPS Delimiters}
\begin{quote}
\small
\ttfamily
TEXT TO CODE:

Survey Response: ``Work/life balance continues to be the biggest challenge, because the amount of work is still elevated and I don't see that changing.''

CODE INFORMATION:

Code: work/life boundaries

Definition: Challenges with maintaining separation between work and personal life, especially during remote work
\end{quote}

\subsection{Numeric Scale Prompts}

\subsubsection{1-5 Scale with XML Delimiters}
\begin{quote}
\small
\ttfamily
\textless TEXT\_TO\_CODE\textgreater

Survey Response: ``Work/life balance continues to be the biggest challenge.''

\textless/TEXT\_TO\_CODE\textgreater

\textless CODE\_INFORMATION\textgreater

Code: work/life boundaries

Definition: Challenges with maintaining separation between work and personal life

\textless/CODE\_INFORMATION\textgreater

\textless TASK\textgreater

Your task is to rate how well the given code applies to the survey response text.

\textless/TASK\textgreater

On a scale from 1-5, where 1 means ``not applicable at all'' and 5 means ``highly applicable'', the score for applying the code in \textless CODE\_INFORMATION\textgreater to the text in \textless TEXT\_TO\_CODE\textgreater is:
\end{quote}

Completion targets: \texttt{`` 1''}, \texttt{`` 2''}, \texttt{`` 3''}, \texttt{`` 4''}, \texttt{`` 5''}

\subsubsection{Reversed 1-5 Scale}
Some conditions test a reversed scale where 1 = highly applicable and 5 = not applicable:

\begin{quote}
\small
\ttfamily
On a scale from 1-5, where 1 means ``highly applicable'' and 5 means ``not applicable at all'', the score for applying this code to this text is:
\end{quote}

\subsection{Qualitative Scale Prompts}

\subsubsection{Intensity Scale}
\begin{quote}
\small
\ttfamily
[Text and Code sections]

\textless TASK\textgreater

Your task is to rate how well the given code applies to the survey response text.

\textless/TASK\textgreater

Use the following scale:
\begin{itemize}[leftmargin=*]
  \item none = the code does not describe the text at all
  \item weak = the code slightly or minimally describes the text
  \item medium = the code moderately describes the text
  \item strong = the code strongly describes the text
  \item perfect = the code perfectly or completely describes the text
\end{itemize}

Using the scale provided above, the intensity of applicability of the code in \textless CODE\_INFORMATION\textgreater to the text in \textless TEXT\_TO\_CODE\textgreater is:
\end{quote}

Completion targets: \texttt{`` none''}, \texttt{`` weak''}, \texttt{`` medium''}, \texttt{`` strong''}, \texttt{`` perfect''}

\subsubsection{Evidence Scale}
\begin{quote}
\small
\ttfamily
Use the following scale:
\begin{itemize}[leftmargin=*]
  \item negligible = there is no evidence that the code applies to the text
  \item weak = there is weak or minimal evidence that the code applies to the text
  \item moderate = there is moderate evidence that the code applies to the text
  \item strong = there is strong or compelling evidence that the code applies to the text
\end{itemize}

Using the scale provided above, the evidence that this code applies to this text is:
\end{quote}

Completion targets: \texttt{`` negligible''}, \texttt{`` weak''}, \texttt{`` moderate''}, \texttt{`` strong''}

\subsubsection{Binary True/False Scale}
\begin{quote}
\small
\ttfamily
Use the following scale:
\begin{itemize}[leftmargin=*]
  \item false = it is false that the code should be applied to the text
  \item true = it is true that the code should be applied to the text
\end{itemize}

Using the scale provided above, the statement `given the context, this code should be applied to this text' is:
\end{quote}

Completion targets: \texttt{`` false''}, \texttt{`` true''}

\subsection{Experimental Factors}

The deductive coding experiments systematically vary:

\begin{itemize}[leftmargin=*]
    \item \textbf{Personas}: None, qualitative researcher, domain expert
    \item \textbf{Scale types}: Numeric (1-5, 1-9) vs. Qualitative (intensity, evidence, binary)
    \item \textbf{Scale anchors}: Different endpoint wordings for numeric scales
    \item \textbf{Scale direction}: Standard (1=low, 5=high) vs. Reversed (1=high, 5=low)
    \item \textbf{Code presentation}: Code+definition, code only, definition only
    \item \textbf{Section delimiters}: XML tags vs. ALL CAPS
    \item \textbf{Context levels}: None, minimal, full deductive coding background
\end{itemize}

\subsection{Complete Example with Persona}
\begin{quote}
\small
\ttfamily
You are an experienced qualitative researcher skilled in systematic coding and analysis of textual data.

\textless TEXT\_TO\_CODE\textgreater

Survey Response: ``Public safety concerns remain paramount, particularly regarding mask mandates and vaccine requirements.''

\textless/TEXT\_TO\_CODE\textgreater

\textless CODE\_INFORMATION\textgreater

Code: public safety (i.e., masks, vaccines, etc.)

Definition: References to public health measures, policies, and safety protocols

\textless/CODE\_INFORMATION\textgreater

\textless TASK\textgreater

Your task is to rate how well the given code applies to the survey response text.

\textless/TASK\textgreater

On a scale from 1-9, where 1 means ``doesn't apply'' and 9 means ``applies perfectly'', the score for applying the code in \textless CODE\_INFORMATION\textgreater to the text in \textless TEXT\_TO\_CODE\textgreater is:
\end{quote}

Completion targets: \texttt{`` 1''} through \texttt{`` 9''}

\section{Ordinal-Scaled Causal Surprisal Curves}
\label{app:ordinal_curves}

\begin{figure*}[t]
    \centering
    \includegraphics[width=\textwidth]{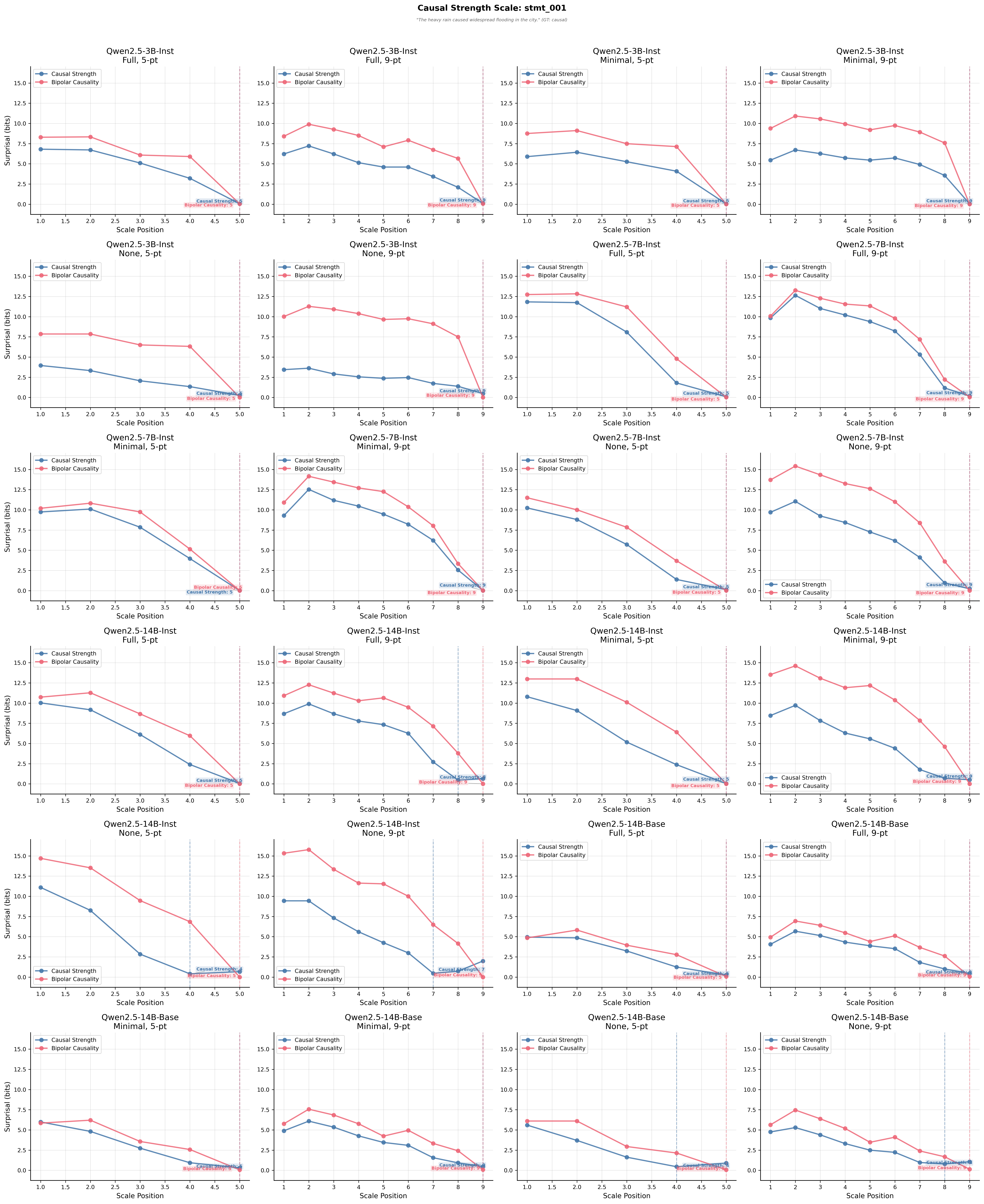}
    \caption{Surprisal curves for the causal statement ``The heavy rainfall caused widespread flooding in the city'' across all model, context, and scale combinations. Each panel shows two scale framings (causal strength and bipolar causality). Monotonically decreasing curves indicate consistent assignment of high causal ratings.}
    \label{fig:causal_statement_surprisal_curves_simple_statement}
\end{figure*}

\begin{figure*}[t]
    \centering
    \includegraphics[width=\textwidth]{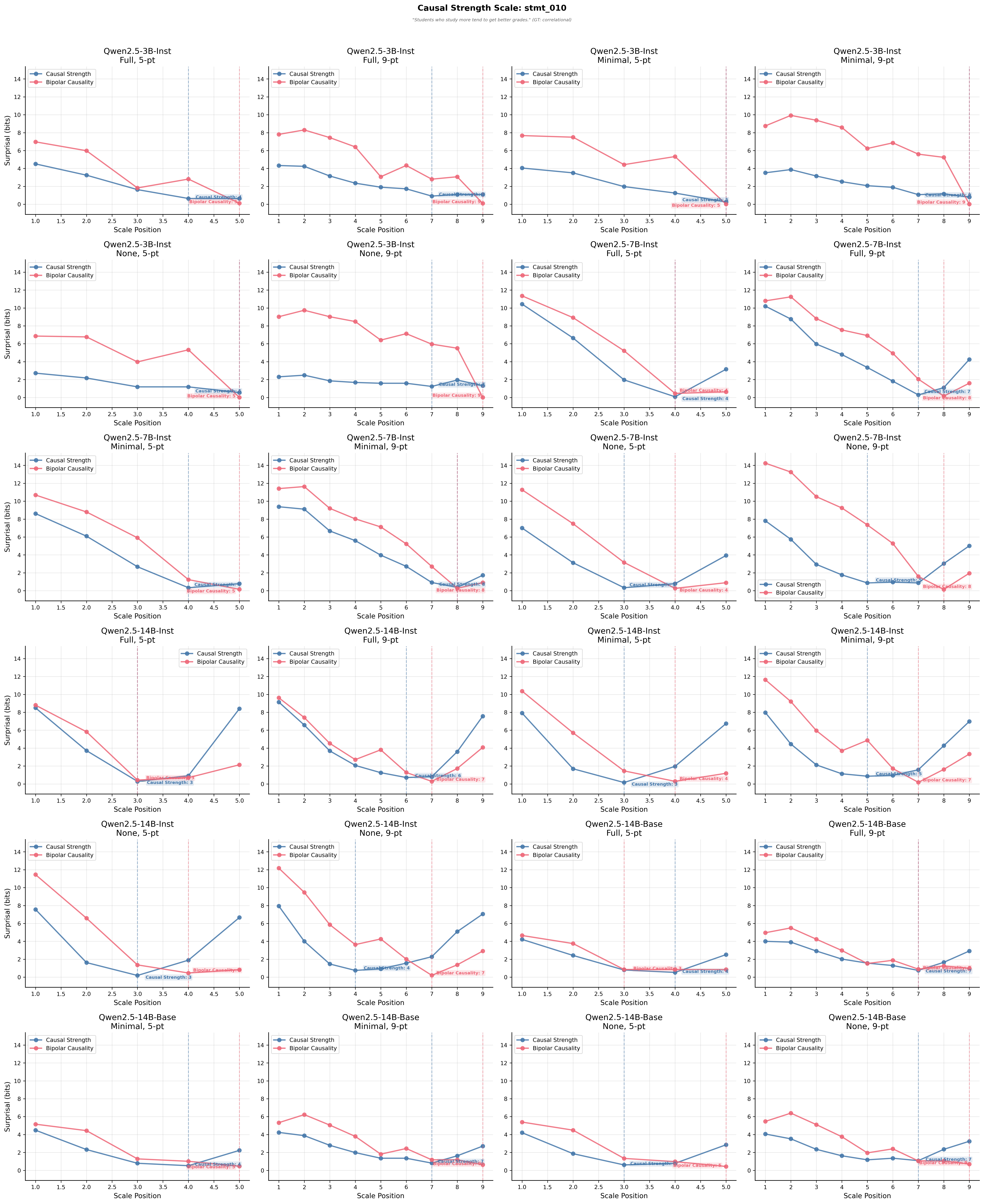}
    \caption{Surprisal curves for the correlational statement ``Students who study more tend to get better grades'' across all model, context, and scale combinations. Compared to the causal statement in Figure~\ref{fig:causal_statement_surprisal_curves_simple_statement}, the curves are flatter and more parabolic, reflecting model uncertainty about the causal content.}
    \label{fig:causal_statement_surprisal_curves_ambiguous_statement}
\end{figure*}

Figures~\ref{fig:causal_statement_surprisal_curves_simple_statement} and~\ref{fig:causal_statement_surprisal_curves_ambiguous_statement} show the full ordinal-scaled surprisal curves for the causal statement experiments discussed in Section~\ref{sec:experiments}. Each figure displays all combinations of model (rows), context level and scale length (columns), with two scale framings (causal strength and bipolar causality) shown as separate lines in each panel.

\end{document}